\documentclass[nohyperref]{article}

\usepackage{microtype}
\usepackage{graphicx}
\usepackage{subfigure}
\usepackage{booktabs}
\usepackage{hyperref}

\usepackage{amsmath,amssymb,graphicx,color,booktabs,bm,relsize,enumitem,multirow,amsthm,epsfig,caption,mathtools,xcolor}
\usepackage{dsfont}
\usepackage{xspace}
\usepackage[export]{adjustbox}
\newsavebox{\measurebox}

\usepackage[accepted]{icml2022}

\usepackage{amsmath}
\usepackage{amssymb}
\usepackage{mathtools}
\usepackage{amsthm}

\usepackage[capitalise,noabbrev,nameinlink]{cleveref}
\creflabelformat{equation}{#2\textup{#1}#3}
\creflabelformat{section}{#2\textup{#1}#3}
\creflabelformat{appendix}{#2\textup{#1}#3}
\creflabelformat{figure}{#2\textup{#1}#3}
\crefname{algocf}{Algorithm}{Algorithms}
\Crefname{algocf}{Algorithm}{Algorithms}

\newcommand{\blap}[1]{\vbox to 0pt{\hbox{#1}\vss}}

\newcommand{\describe}[3][0pt]{\hspace*{.12em}\underbracket[0.5pt][1pt]{#2\hspace*{#1}}_\text{#3}}

\theoremstyle{plain}

\theoremstyle{definition}

\theoremstyle{remark}

\usepackage[textsize=tiny]{todonotes}

\icmltitlerunning{
Reinforcement Learning with Action-Free Pre-Training from Videos
}
\begin{document}

\twocolumn[
\icmltitle{
Reinforcement Learning with Action-Free Pre-Training from Videos
}
\icmlsetsymbol{equal}{*}

\begin{icmlauthorlist}
\icmlauthor{Younggyo Seo}{kaist,visit}
\icmlauthor{Kimin Lee}{ucb,google}
\icmlauthor{Stephen James}{ucb}
\icmlauthor{Pieter Abbeel}{ucb}
\end{icmlauthorlist}

\icmlaffiliation{kaist}{KAIST}
\icmlaffiliation{ucb}{UC Berkeley}
\icmlaffiliation{visit}{Work done while visiting UC Berkeley}
\icmlaffiliation{google}{Now at Google Research}

\icmlcorrespondingauthor{Younggyo Seo}{younggyo.seo@kaist.ac.kr}

\icmlkeywords{Machine Learning, Reinforcement Learning, ICML}

\vskip 0.3in
]

\newcommand{\ALG}{APV\xspace}
\newcommand{\ALGname}{\textbf{A}ction-Free \textbf{P}re-training from \textbf{V}ideos\xspace}

\printAffiliationsAndNotice{}

\begin{abstract}
Recent unsupervised pre-training methods have shown to be effective on language and vision domains by learning useful representations for multiple downstream tasks.
In this paper, we investigate if such unsupervised pre-training methods can also be effective for vision-based reinforcement learning (RL).
To this end, we introduce a framework that learns representations useful for understanding the dynamics via generative pre-training on videos.
Our framework consists of two phases: we pre-train an action-free latent video prediction model, and then utilize the pre-trained representations for efficiently learning action-conditional world models on unseen environments.
To incorporate additional action inputs during fine-tuning, we introduce a new architecture that stacks an action-conditional latent prediction model on top of the pre-trained action-free prediction model.
Moreover, for better exploration, we propose a video-based intrinsic bonus that leverages pre-trained representations.
We demonstrate that our framework significantly improves both final performances and sample-efficiency of vision-based RL in a variety of manipulation and locomotion tasks.
Code is available at \url{https://github.com/younggyoseo/apv}.
\end{abstract}

\section{Introduction}
\label{sec:introduction}
Deep reinforcement learning (RL) has made significant advance in solving various sequential decision-making problems~\citep{mnih2015human,levine2016end,silver2017mastering,vinyals2019grandmaster,berner2019dota,akkaya2019solving,kalashnikov2021mt}.
However, existing RL methods often start learning \textit{tabula rasa} without any prior knowledge of the world, therefore requiring a large amount of environment interaction for learning meaningful behaviors. By contrast, within the computer vision (CV) and natural language processing (NLP) domains, recent unsupervised pre-training approaches have shown to be effective by leveraging the pre-trained representations for fine-tuning in downstream tasks~\citep{mikolov2013efficient,pennington2014glove,noroozi2016unsupervised,gidaris2018unsupervised,devlin2018bert,radford2018improving,he2020momentum}.

\begin{figure}
    \centering
    \includegraphics[width=0.48\textwidth]{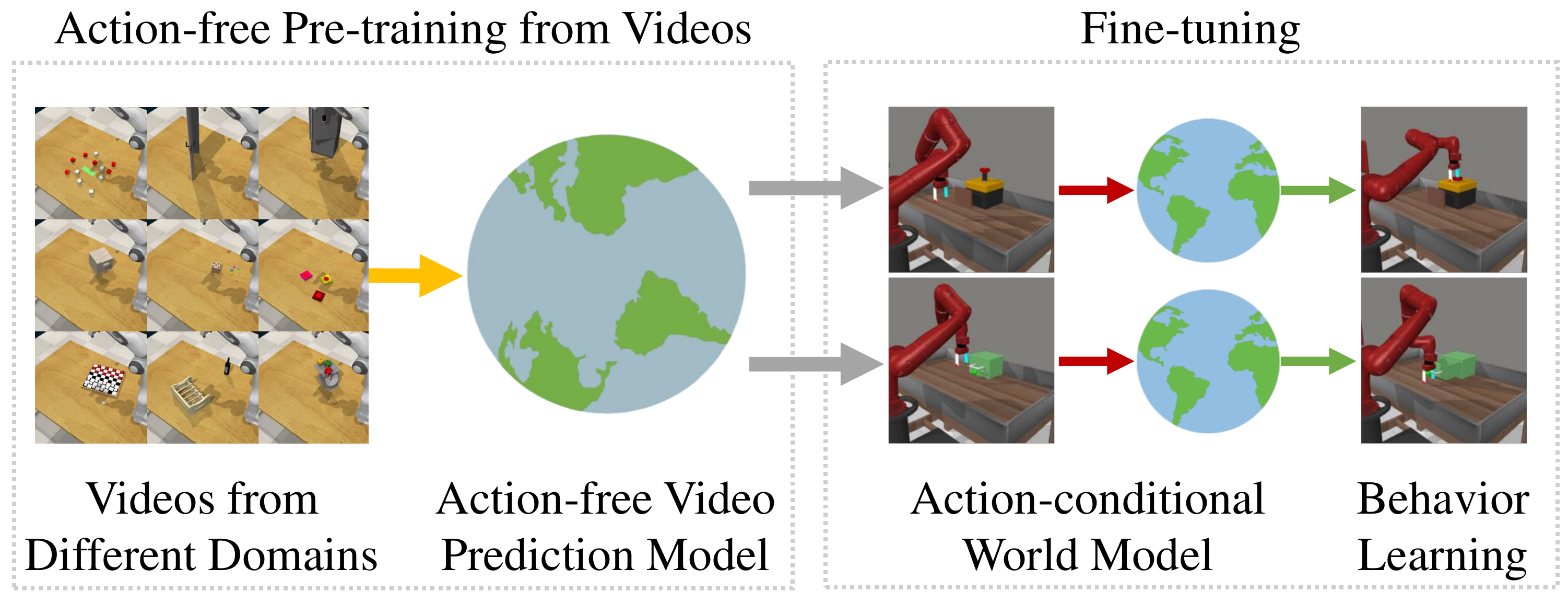}
    \vspace{-0.25in}
    \caption{We pre-train an action-free latent video prediction model using videos from different domains (left), and then fine-tune the pre-trained model on target domains (right).}
    \label{fig:concept_figure}
\end{figure}

Recent works have shown promise in adopting such \textit{pre-training and fine-tuning} paradigm to vision-based RL, by demonstrating that representations pre-trained with various unsupervised representation learning schemes can be effective in downstream tasks~\citep{finn2016deep,dwibedi2018learning,zhan2020framework,laskin2020curl,stooke2021decoupling,schwarzer2021pretraining}.
Notably, \citet{stooke2021decoupling} show that weight initialization with contrastively pre-trained representations leads to performance improvement.
These works, however, mostly focus on the setup where pre-training datasets are collected in the target domains, or in different domains but with very similar visuals.
Instead, we would like to leverage videos from diverse domains for pre-training, and transfer the pre-trained representations for solving newly encountered domains.

In this paper, we present \ALG: \ALGname, a novel framework that performs generative pre-training on videos for improving the sample-efficiency of vision-based RL.
Since our goal is to learn the representations that can be transferred to various downstream tasks from readily available videos, our framework do not require the videos to be collected in the same domain of the downstream tasks, and also do not assume the datasets contain action information.
Summarized in \cref{fig:concept_figure}, our framework comprises two phases: we first pre-train an action-free latent video prediction model to learn useful representations from videos, then fine-tune the pre-trained model for learning action-conditional world models on downstream tasks.
To leverage the fine-tuned world models for behavior learning, we build \ALG on top of DreamerV2~\citep{hafner2020mastering}.

The key ingredients of \ALG are as follows:
\begin{itemize} [topsep=0pt,itemsep=0pt,leftmargin=*]
\item [$\bullet$] \textbf{Action-free pre-training from videos}: To capture rich dynamics information from diverse videos, we pre-train an action-free latent video prediction model.
We find that the representations from the pre-trained model can be transferred to various downstream tasks.
\item [$\bullet$] \textbf{Stacked latent prediction model}: To incorporate additional action inputs during fine-tuning,
we introduce a new architecture that stacks an action-conditional latent dynamics model on top of the action-free model.
\item [$\bullet$] \textbf{Video-based intrinsic bonus}: For better exploration, we propose an intrinsic bonus that utilizes video representations from the action-free model.
Since the pre-trained representations contain information useful for understanding dynamics of environments,
our intrinsic bonus effectively encourages agents to learn diverse behaviors.
\end{itemize}

In our experiments, we pre-train the action-free prediction model using 4950 videos collected on 99 manipulation tasks from RLBench~\citep{james2020rlbench} and fine-tune the pre-trained model on a range of manipulation tasks from Meta-world~\citep{yu2020meta}.
Despite a big domain gap between RLBench and Meta-world, we demonstrate that \ALG significantly outperforms DreamerV2.
For example, \ALG achieves the aggregate success rate of 95.4\% on six manipulation tasks, while DreamerV2 achieves 67.9\%.
Moreover, we show that RLBench pre-trained representations can also be effective in learning locomotion tasks from DeepMind Control Suite~\citep{tassa2020dm_control}, where both the visuals and objectives significantly differ from RLBench videos.

\section{Related Work}
\label{sec:related_work}

\paragraph{Unsupervised representation learning for CV and NLP.}
Recently, unsupervised representation learning methods have been actively studied in the domain of CV.
Various representation learning methods, including reconstruction~\cite{he2021masked}, rotation~\citep{gidaris2018unsupervised}, solving zigsaw puzzles~\citep{noroozi2016unsupervised}, and contrastive learning~\citep{he2020momentum,chen2020simple}, have reduced the gap with supervised pre-training with labels.
In the domain of NLP, unsupervised representation learning has been successfully applied to training language models with generalization ability~\citep{devlin2018bert,radford2018improving,yang2019xlnet}.
Notably, \citet{devlin2018bert} and \citet{radford2018improving} trained large transformer networks~\citep{vaswani2017attention} with masked token prediction and generative pre-training, respectively, and showed that pre-trained models can be effectively fine-tuned on downstream tasks.
In this work, we demonstrate that unsupervised pre-training can also be effective for vision-based RL.

\paragraph{Unsupervised representation learning for RL.}
Unsupervised representation learning for RL has also been studied to improve the sample-efficiency of RL algorithms. Notably, \citet{jaderberg2016reinforcement} showed that optimizing auxiliary unsupervised losses can improve the performance of RL agents.
This has been followed by a series of works which demonstrated the effectiveness of various unsupervised learning objectives, including world-model learning~\citep{hafner2019learning,hafner2020mastering}, reconstruction~\citep{yarats2019improving}, future representation prediction~\citep{gelada2019deepmdp,schwarzer2020data}, bisimulation~\citep{castro2020scalable,zhang2020learning}, and contrastive learning~\citep{oord2018representation,anand2019unsupervised,mazoure2020deep,srinivas2020curl}.
While these works optimize auxiliary unsupervised objectives to accelerate the training of RL agents, we instead aim to pre-train representations as in CV and NLP domains.

There have been several approaches to perform unsupervised pre-training for RL~\citep{finn2016deep,dwibedi2018learning,zhan2020framework,srinivas2020curl,stooke2021decoupling,schwarzer2021pretraining}. In particular, \citet{schwarzer2021pretraining} proposed several self-supervised learning objectives that rely on actions, but assume access to action information of downstream tasks which may not be available in practice.
The work closest to ours is \citet{stooke2021decoupling}, which demonstrated that the representations contrastively pre-trained without actions and rewards can be effective on unseen downstream tasks but with very similar visuals.
In this work, we instead develop a framework that leverages action-free videos from diverse domains with different visuals and embodiments for pre-training.
Concurrent to our work, \citet{xiao2022masked} showed that pre-training a visual encoder on videos can be effective for downstream RL tasks.
We instead pre-train a video prediction model instead of a visual encoder that operates on a single image.

\paragraph{Behavior learning with videos.}
Video datasets have also been utilized for behavior learning in various ways~\citep{peng2018sfv,torabi2018generative,aytar2018playing,liu2018imitation,sermanet2018time,edwards2019imitating,schmeckpeper2020learning,schmeckpeper2020reinforcement,chang2020semantic,chen2021learning,zakka2022xirl}.
\citet{aytar2018playing} solved hard exploration tasks on Atari benchmark by designing an imitation reward based on YouTube videos, and \citet{peng2018sfv} proposed to learn physical skills from human demonstration videos by extracting reference motions and training an RL agent that imitates the extracted motions.
Our work differs in that we utilize videos for pre-training representations, instead of learning behaviors from videos.

We provide more discussion on related fields in \cref{appendix:extended_related_work}.

\section{Method}
\label{sec:framework}
We formulate a vision-based control task as a partially observable Markov decision process (POMDP), which is defined as a tuple $\left( \mathcal{O}, \mathcal{A}, p, r, \gamma\right)$.
Here, $\mathcal{O}$ is the high-dimensional observation space, $\mathcal{A}$ is the action space, $p\left(o_{t}|o_{< t}, a_{< t}\right)$ is the transition dynamics, 
$r$ is the reward function that maps previous observations and actions to a reward $r_{t} = r\left(o_{\leq t}, a_{< t}\right)$,
and $\gamma \in [0,1)$ is the discount factor.
The goal of RL is to learn an agent that behaves to maximize the expected sum of rewards $\mathbb{E}_{p}\left[\sum^{T}_{t=1} \gamma^{t-1} r_{t}\right]$.

\subsection{Action-free Pre-training from Videos}
\label{sec:action_free_pretraining}
\begin{figure}[t!] \centering
\includegraphics[width=.48\textwidth]{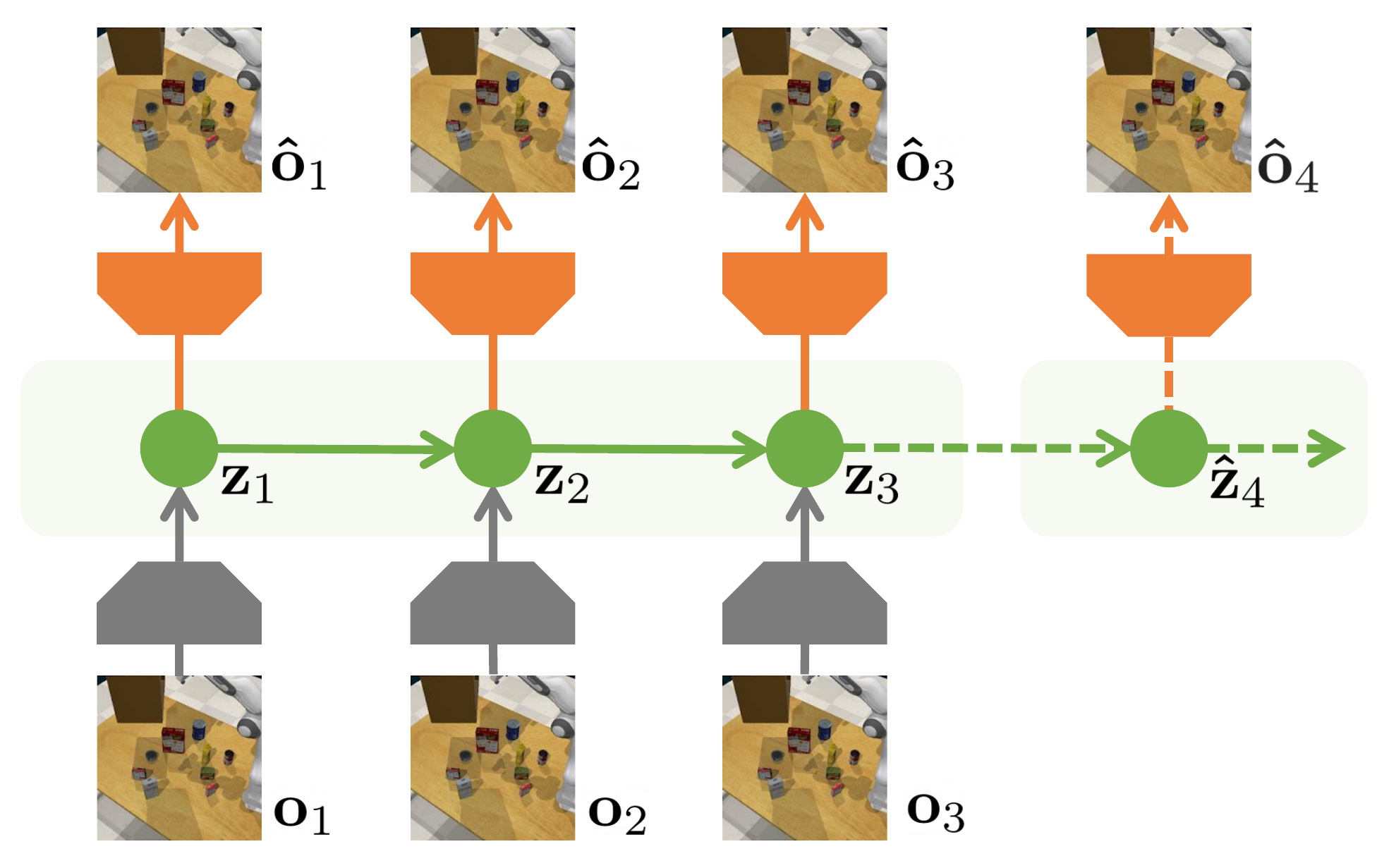}
\vspace{-0.35in}
\caption{Illustration of action-free latent video prediction model.
The model is trained to capture visual and dynamics information from action-free videos by reconstructing image observations.
At inference time, the transition model is used to predict future states in the latent space without conditioning on predicted frames.}
\label{fig:method_action_free_pretraining}
\vspace{-0.0in}
\end{figure}

In order to pre-train representations from action-free videos, we first learn a latent video prediction model, which is an action-free variant of a latent dynamics model~\citep{hafner2019learning}.
Unlike autoregressive video prediction models that predict a next frame and utilize it as an input for the following prediction,
the model instead operates on the latent space~\citep{zhang2019solar,hafner2019learning,franceschi2020stochastic}.
Specifically, the model consists of three main components: (i) the representation model that encodes observations $o_{t}$ to a model state $z_{t}$ with Markovian transitions, (ii) the transition model that predicts future model states $\hat{z}_{t}$ without access to the observation, and (iii) the image decoder that reconstructs image observations $\hat{o}_{t}$.
The model can be summarized as follow (see~\cref{fig:method_action_free_pretraining}):
\begin{gather}
\begin{aligned}
&\text{Representation model:} &&z_t\sim q_\phi(z_{t} \,|\,z_{t-1},o_{t}) \\
&\text{Transition model:} &&\hat{z}_t\sim p_\phi(\hat{z}_{t} \,|\,z_{t-1}) \\
&\text{Image decoder:} &&\hat{o}_t\sim p_\phi(\hat{o}_{t} \,|\,z_{t})
\label{eq:action_free_model}
\end{aligned}
\end{gather}
We train the model to reconstruct image observations, and to make the prediction from the representation model and transition model be close to each other.
All model parameters $\phi$ are jointly optimized by minimizing the negative variational lower bound (ELBO;~\citealt{kingma2013auto}):
\begin{align}
    &\mathcal{L}(\phi) \doteq \;\mathbb{E}_{q_{\phi}\left(z_{1:T}\,|\,o_{1:T}\right)}\Big[
    \textstyle\sum_{t=1}^{T} \Big(
    \describe{-\ln p_{\phi}(o_{t}\,|\,z_{t})}{image log loss} \nonumber \\
    &\quad\quad\quad\describe{+\beta_{z}\,\text{KL}\left[q_{\phi}(z_{t}\,|\,z_{t-1},o_{t}) \,\Vert\,p_{\phi}(\hat{z}_{t}\,|\,z_{t-1}) \right]}{action-free KL loss}
    \Big)\Big], \label{eq:action_free_model_objective}
\end{align}
where $\beta_{z}$ is a scale hyperparameter and $T$ is the length of training sequences in a minibatch.
Since the transition model does not condition on observations, it allows us to efficiently predict future states in the latent space without needing to predict future images using the image decoder at inference time.
We implement the transition model as an action-free recurrent state-space model (RSSM;~\citealt{hafner2019learning}), which consists of both deterministic and stochastic components, and the representation model by combining the action-free RSSM with an image encoder.
We refer to \cref{appendix:formulation_rssm} for a more detailed formulation.

\begin{figure*}[t!] \centering
\subfigure[Stacked latent prediction model]
{
\includegraphics[width=0.55545455\textwidth]{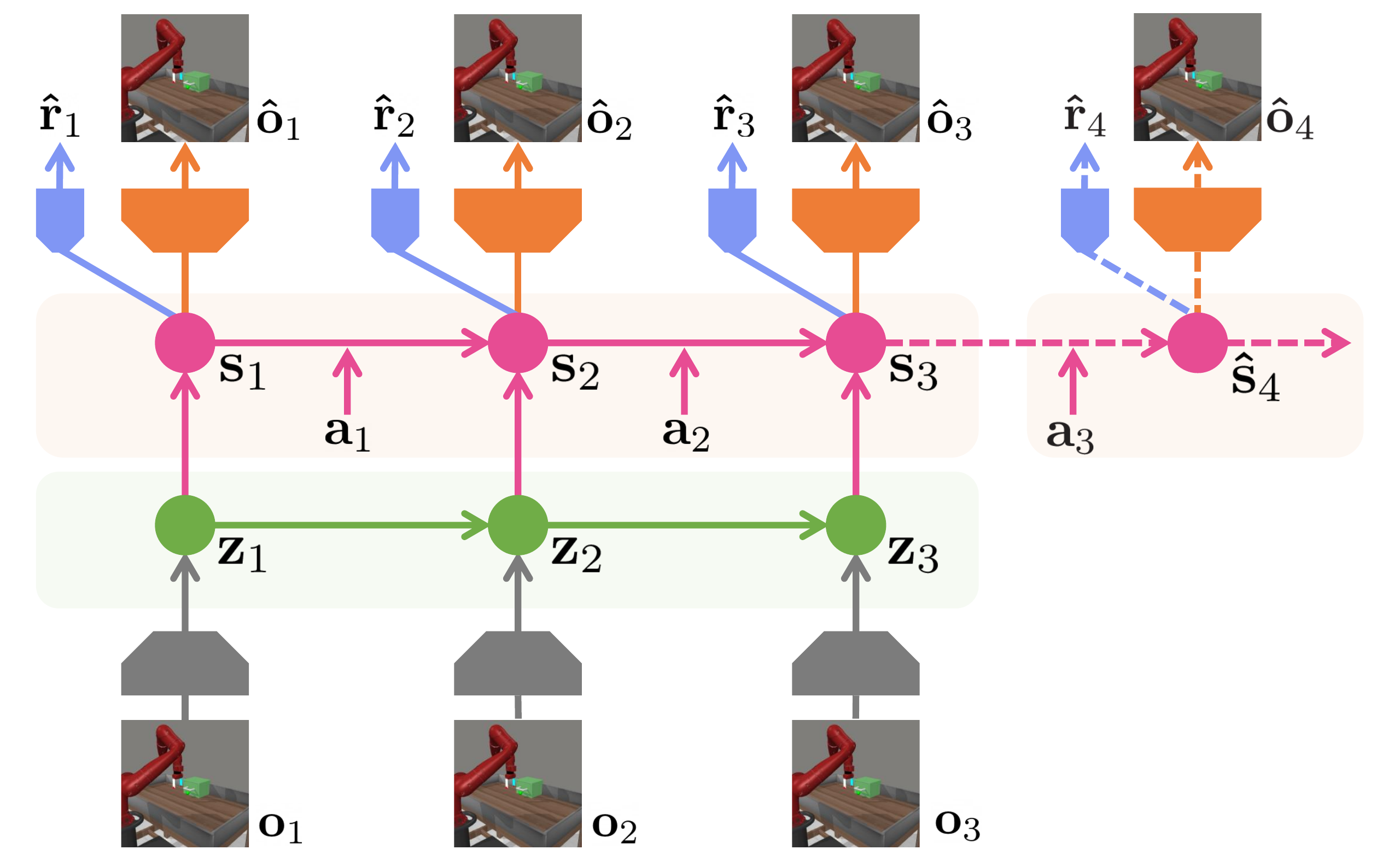}
\label{fig:method_stack}}
\subfigure[Video-based intrinsic bonus]
{
\includegraphics[width=0.38454545\textwidth]{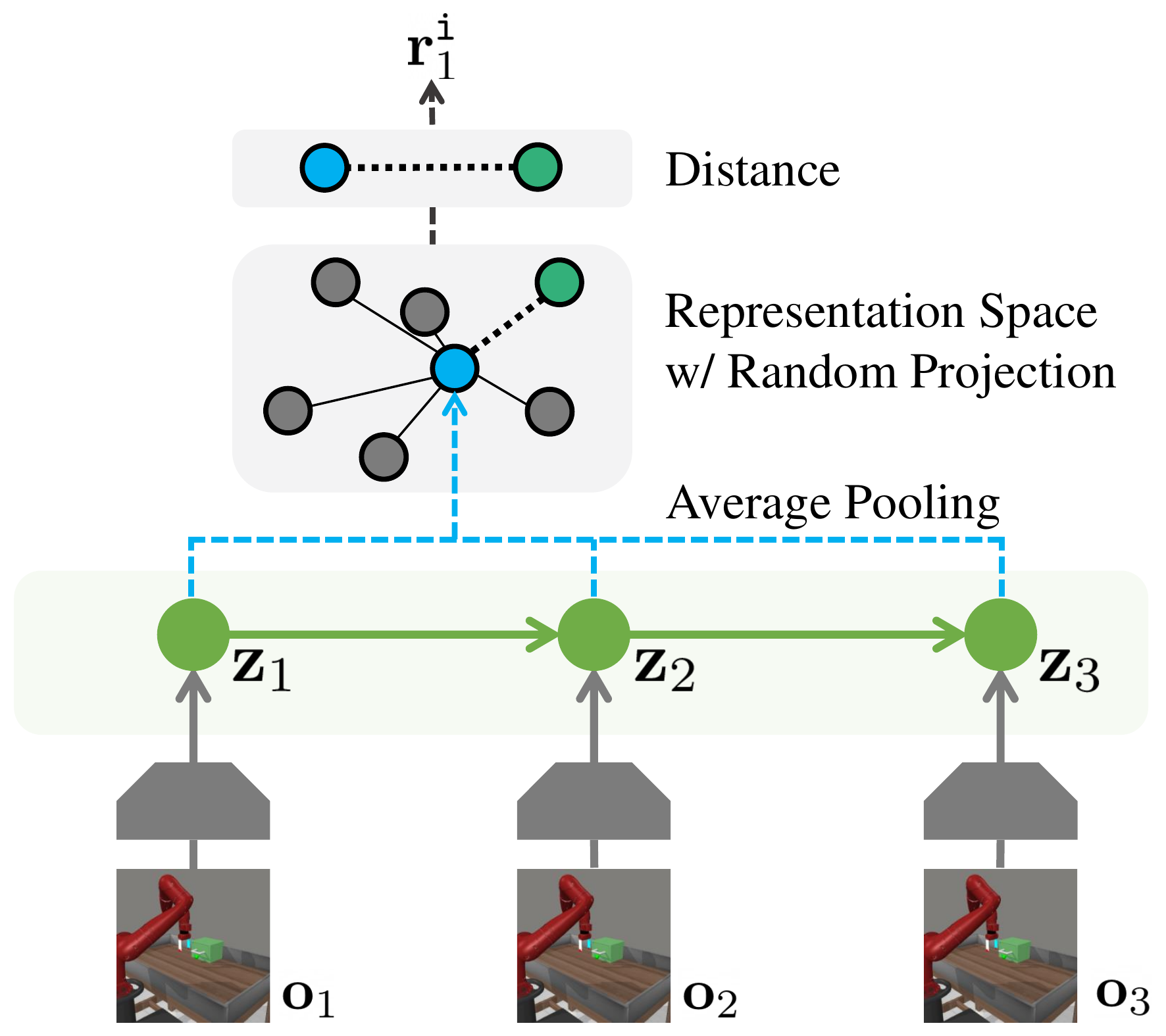}
\label{fig:method_intrinsic}}
\caption{
Illustration of our framework.
(a) We stack an action-conditional prediction model on top of the pre-trained action-free prediction model. At inference time, the transition model in the action-conditional model is used to predict future states in the latent space conditioned on future potential actions.
(b) To compute the intrinsic bonus, we first average pool a sequence of model states from the action-free prediction model, and apply random projection to reduce the dimension of representations while preserving distances. The intrinsic bonus for each observation is computed as the distance in the representation space to its $k$-nearest neighbor in samples from a replay buffer.}
\label{fig:method_finetuning}
\end{figure*}

\subsection{Stacked Latent Prediction Model}
\label{sec:stacked_rssm}
Once we pre-train the action-free prediction model, we fine-tune it into an action-conditional prediction model that can be used for solving various visual control tasks.
Since actions and rewards, which provide more information about target tasks, are available during fine-tuning, it motivates incorporating them into the model.
One na\"ive approach would be to initialize the action-conditional prediction model with the action-free model, and learn a reward predictor on top of it.
But we find this fine-tuning scheme rapidly erases the useful knowledge in pre-trained models (see~\cref{fig:metaworld_naive_finetuning} for supporting results).
To effectively utilize the pre-trained representations, we introduce a new architecture that stacks an action-conditional prediction model on top of the action-free model as below (see~\cref{fig:method_stack}):
\begin{align}
&\text{\textbf{Action-free}} \nonumber\\[-0.04in]
\raisebox{1.95ex}{\llap{\blap{\ensuremath{ \hspace{0.1ex} \begin{cases} \hphantom{R} \\ \hphantom{T} \end{cases} \hspace*{-4ex}
}}}}
&\text{Representation model:} &&z_t\sim q_\phi(z_{t} \,|\,z_{t-1},o_{t}) \nonumber\\
&\text{Transition model:} &&\hat{z}_t\sim p_\phi(\hat{z}_{t} \,|\,z_{t-1}) \nonumber\\
&\text{\textbf{Action-conditional}} \nonumber\\[-0.05in]
\raisebox{1.95ex}{\llap{\blap{\ensuremath{ \hspace{0.1ex} \begin{cases} \hphantom{R} \\ \hphantom{T} \end{cases} \hspace*{-4ex}
}}}}
&\text{Representation model:} &&s_t\sim q_\theta(s_{t} \,|\,s_{t-1},a_{t-1}, z_{t}) \nonumber\\
&\text{Transition model:} &&\hat{s}_t\sim p_\theta(\hat{s}_{t} \,|\,s_{t-1}, a_{t-1}) \nonumber\\[0.02in]
&\text{Image decoder:} &&\hat{o}_t\sim p_\theta(\hat{o}_{t} \,|\,s_{t}) \nonumber\\
&\text{Reward predictor:} &&\hat{r}_t\sim p_\theta(\hat{r}_{t} \,|\,s_{t}),
\label{eq:stacked_world_model}
\end{align}
which is optimized by minimizing the following objective:
\begin{align}
    &\mathcal{L}(\phi,\theta) \doteq  \mathbb{E}_{q_{\theta}\left(s_{1:T}|a_{1:T},z_{1:T}\right),q_{\phi}\left(z_{1:T}\,|\,o_{1:T}\right)}\Big[\nonumber \\
    &\textstyle\sum_{t=1}^{T} \Big(
    \describe{-\ln p_{\theta}(o_{t}|s_{t})}{image log loss}
    \describe{-\ln p_{\theta}(r_{t}|s_{t})}{reward log loss} \nonumber \\
    &\quad\;\; \describe{+ \beta_{z}\,\text{KL}\left[ q_{\phi}(z_{t}|z_{t-1},o_{t}) \,\Vert\,p_{\phi}(\hat{z}_{t}|z_{t-1}) \right]}{action-free KL loss} \label{eq:stacked_world_model_objective}\\
    &\quad\;\; \describe{+ \beta\,\text{KL}\left[ q_{\theta}(s_{t}|s_{t-1},a_{t-1},z_{t}) \,\Vert\,  p_{\theta}(\hat{s}_{t}|s_{t-1},a_{t-1}) \right]}{action-conditional KL loss}
    \Big)\Big],\nonumber
\end{align}
where $\beta$ is a scale hyperparameter.
Here, we note that we initialize the image decoder $p_{\theta}(o|s_{t})$ with the pre-trained image decoder $p_{\phi}(o|z_{t})$.
We implement the transition model of action-conditional prediction model as RSSM, and the representation model as RSSM with dense layers that receive the model states of the action-free model as inputs.
We refer to \cref{appendix:formulation_rssm} for a more detailed formulation.
In our experiments, we use $\beta_{z} = 0$ during fine-tuning, and only utilize the action-conditional RSSM for future imagination.

\subsection{Video-based Intrinsic Bonus}
\label{sec:video_based_intrinsic_bonus}
It has been observed that good representations are crucial for efficient exploration in environments with high-dimensional observations~\citep{laskin2021urlb}.
To utilize useful information captured in the pre-trained representations for exploration, we propose a video-based intrinsic bonus.
Our main idea is to increase the diversity of visited trajectories by utilizing it as an intrinsic bonus.
Specifically, given a sequence of model states from the action-free prediction model $z_{t: t+\tau}$,
we apply average pooling across the sequence dimension to obtain a trajectory representation $y_{t} = \text{Avg}(z_{t: t+\tau})$.
Then, we utilize the distance of $y_{t}$ to its $k$-nearest neighbor in samples from a replay buffer as a metric for measuring the diversity of trajectories.
To summarize, our intrinsic bonus is defined as below (see~\cref{fig:method_intrinsic} for illustration):
\begin{align}
    r^{\tt{int}}_{t} \doteq ||\psi(y_{t}) - \psi(y^{k}_{t})||_{2},
    \label{eqn:intrinsic_bonus}
\end{align}
where $\psi$ is a random projection~\citep{bingham2001random} that maps the model state to a low-dimensional representation for compute-efficient distance computation, and $y_{t}^{k}$ is a $k$-nearest neighbor of $y_{t}$ in a minibatch.
By explicitly encouraging the agents to visit more diverse trajectories instead of single states~\citep{pathak2017curiosity,burda2018exploration,pathak2019self,liu2021behavior},
it effectively encourages the agents to explore environments in a more long term manner, and thus learn more diverse behaviors.
Then, the reward predictor is trained to predict the sum of $r_{t}$ and $r_{t}^{\tt{int}}$ as below:
\begin{align}
    &\mathcal{L}^{\tt{APV}}(\phi,\theta) \doteq  \mathbb{E}_{q_{\theta}\left(s_{1:T}|a_{1:T},z_{1:T}\right),q_{\phi}\left(z_{1:T}\,|\,o_{1:T}\right)}\Big[\nonumber \\
    &\textstyle\sum_{t=1}^{T} \Big(
    \describe{-\ln p_{\theta}(o_{t}|s_{t})}{image log loss}
    \describe{-\ln p_{\theta}(r_{t}+ \lambda r_{t}^{\tt{int}}\,|\,h_{t},z_{t})}{\ALG reward log loss} \nonumber \\
    &\quad\;\; \describe{+ \beta_{z}\,\text{KL}\left[ q_{\phi}(z_{t}|z_{t-1},o_{t}) \,\Vert\,p_{\phi}(\hat{z}_{t}|z_{t-1}) \right]}{action-free KL loss} \label{eq:apv_finetuning_objective}\\
    &\quad\;\; \describe{+ \beta\,\text{KL}\left[ q_{\theta}(s_{t}|s_{t-1},a_{t-1},z_{t}) \,\Vert\,  p_{\theta}(\hat{s}_{t}|s_{t-1},a_{t-1}) \right]}{action-conditional KL loss}
    \Big)\Big],\nonumber
\end{align}
where $\lambda$ is a hyperparameter that adjusts the tradeoff between exploitation and exploration.
We find the intrinsic bonus provides large gains when combined with pre-training, as the pre-trained representations already contain useful representation from the beginning of the fine-tuning (see~\cref{fig:metaworld_tsne} for supporting results).
In our experiments, we utilize a sliding window of size $\tau$ for constructing a set of $\{y_{t}\}$ from trajectories in a minibatch, then compute the intrinsic bonus using them.
For behavior learning, we utilize the actor-critic learning scheme of DreamerV2 \citep{hafner2020mastering} that learns values with imagined rewards from future imaginary states and a policy that maximizes the values (see~\cref{appendix:behavior_learning} for details).
We also summarize the difference of \ALG to DreamerV2 in~\cref{appendix:dreamerv2_difference}.
\clearpage

\section{Experiments}
\label{sec:experiments}
We designed our experiments to investigate the following:
\begin{itemize}[topsep=0.0pt,itemsep=0.0pt,leftmargin=10pt]
\item [$\bullet$] Can \ALG improve the sample-efficiency of vision-based RL in robotic manipulation tasks by performing action-free pre-training on videos from different domains?
\item [$\bullet$] Can representations pre-trained on videos from manipulation tasks transfer to locomotion tasks?
\item [$\bullet$] How does \ALG compare to a na\"ive fine-tuning scheme?
\item [$\bullet$] What is the contribution of each of the proposed techniques in \ALG?
\item [$\bullet$] How does pre-trained representations qualitatively differ from the randomly initialized representations?
\item [$\bullet$] How does \ALG perform when additional in-domain videos or real-world natural videos are available?
\end{itemize}
Following \citet{agarwal2021deep}, we report the interquartile mean with bootstrap confidence interval (CI) and stratified bootstrap CI for results on individual tasks and aggregate results, respectively, across 8 runs for each task.
Source codes and other resources are available at \url{https://github.com/younggyoseo/apv}.

\subsection{Experimental Setup}
\label{sec:experimental_setup}
\begin{figure}[t]
    \centering
    \includegraphics[width=0.48\textwidth]{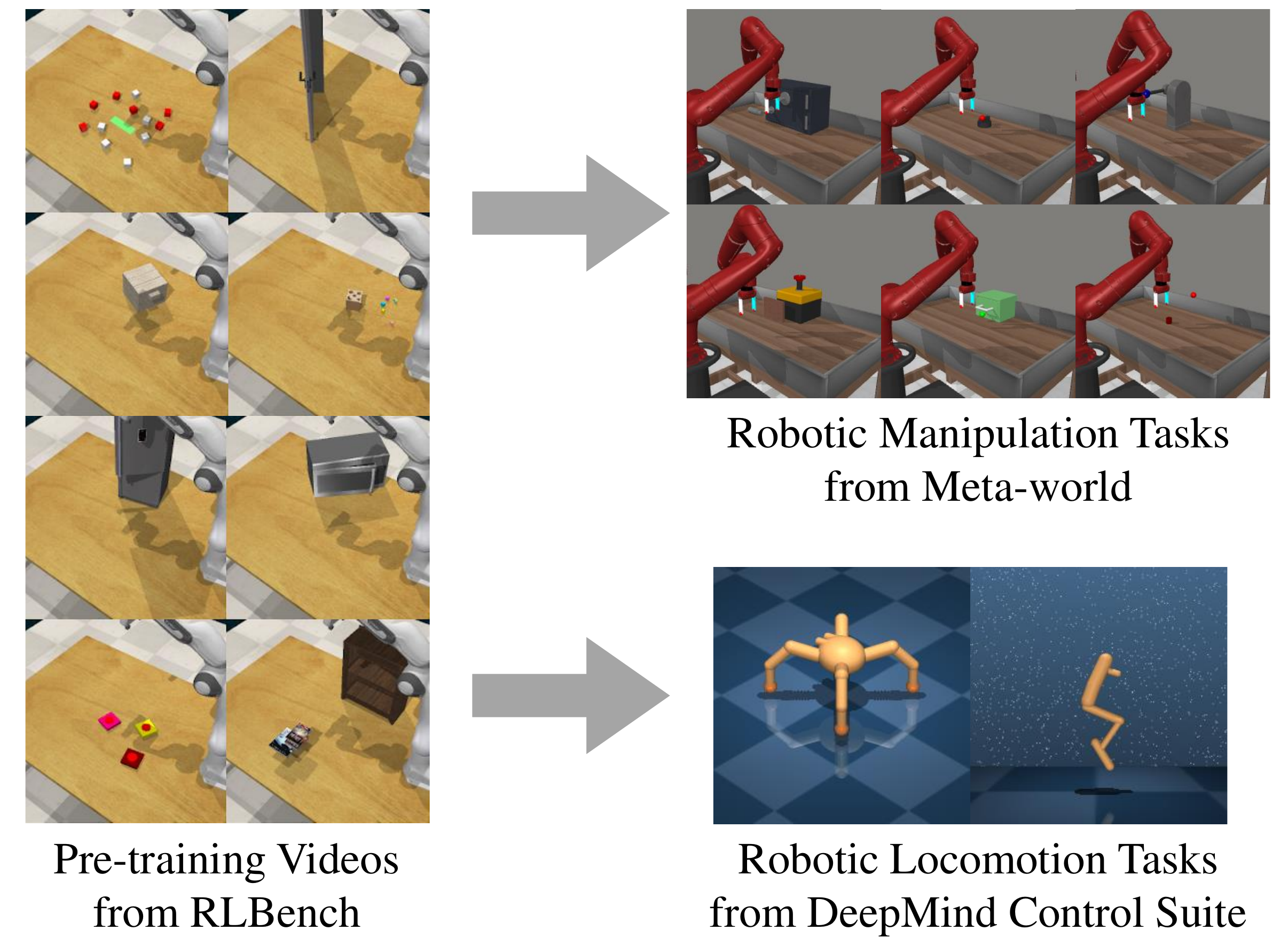}
    \vspace{-0.3in}
    \caption{
    Illustration of experimental setups in our experiments with examples of image observations from environments.
    One can see that visuals in pre-training videos are notably different from the visuals in downstream manipulation and locomotion tasks.}
    \vspace{-0.1in}
    \label{fig:setup}
\end{figure}

\paragraph{Meta-world experiments.} We first evaluate \ALG on various vision-based robotic manipulation tasks from Meta-world~\citep{yu2020meta}.
In all manipulation tasks, the episode length is 500 steps without any action repeat, action dimension is 4, and reward ranges from 0 to 10.
To evaluate the ability of \ALG to learn useful representations from different domains, we use videos collected in robotic manipulation tasks from RLBench~\citep{james2020rlbench} as pre-training data (see~\cref{fig:setup}).\footnote{In this work, we do not consider a setup where we perform pre-training on Meta-world videos and fine-tuning for solving RLBench manipulation tasks, as existing RL algorithms struggle to solve challenging, sparsely-rewarded RLBench tasks.}
Specifically, we collect 10 demonstrations rendered with 5 camera views in 99 tasks from RLBench; giving a total of 4950 videos.
We then train the action-free video prediction model by minimizing the objective in \cref{eq:action_free_model_objective} for 600K gradient steps.
For downstream tasks, we fine-tune the model by minimizing the objective in~\cref{eq:apv_finetuning_objective} for 250K environment steps, i.e., 500 episodes.

\begin{figure*} [t!] \centering
\includegraphics[width=0.99\textwidth]{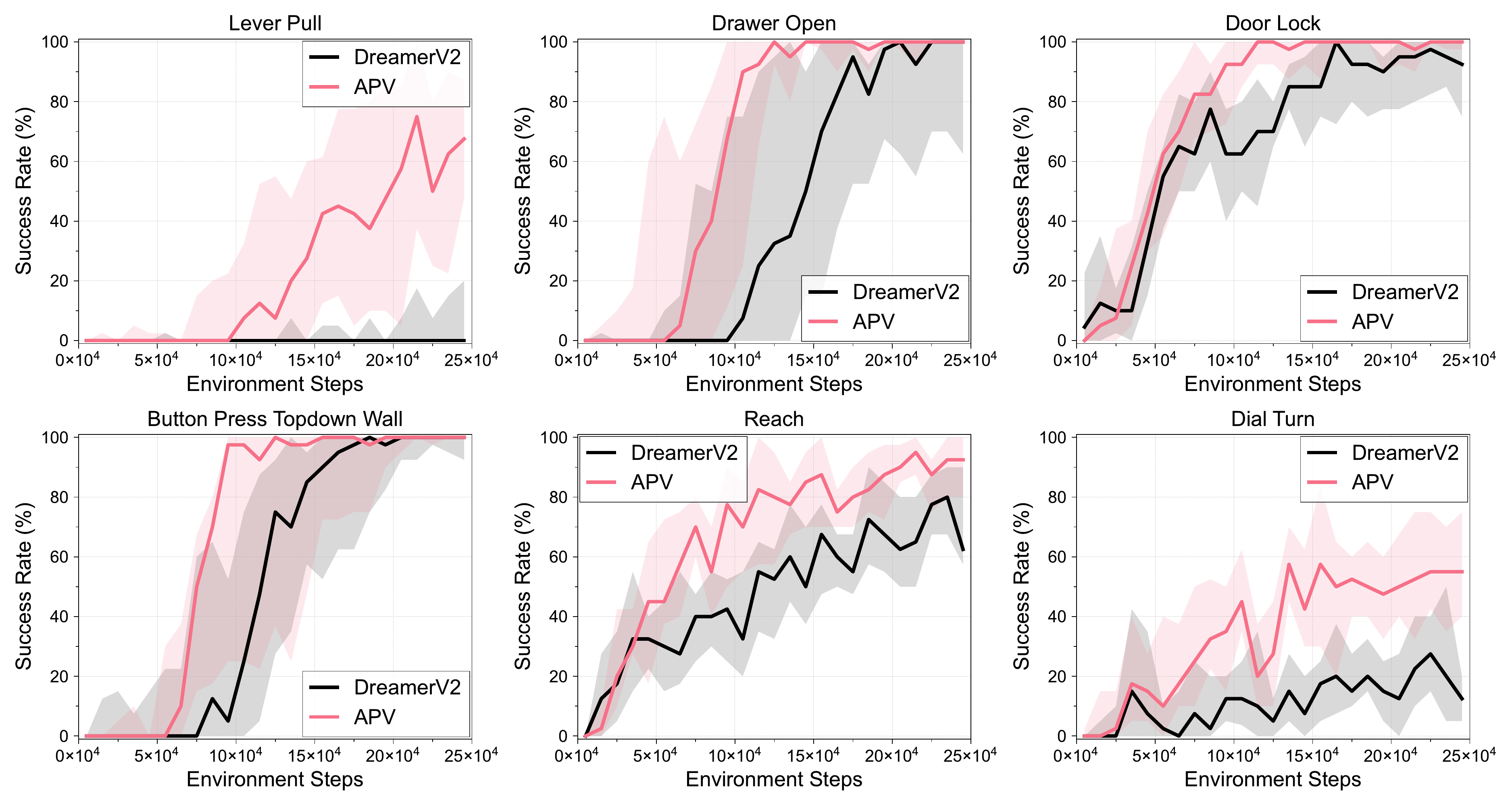}
\vspace{-0.2in}
\caption{
Learning curves on manipulation tasks from Meta-world as measured on the success rate.
\ALG with generative pre-training on videos collected in manipulation tasks from RLBench consistently outperforms DreamerV2 in terms of sample-efficiency.
The solid line and shaded regions represent the interquartile mean and bootstrap confidence intervals, respectively, across eight runs.}
\label{fig:metaworld_performance}
\end{figure*}

\begin{figure}[t]
    \centering
    \includegraphics[width=0.4\textwidth]{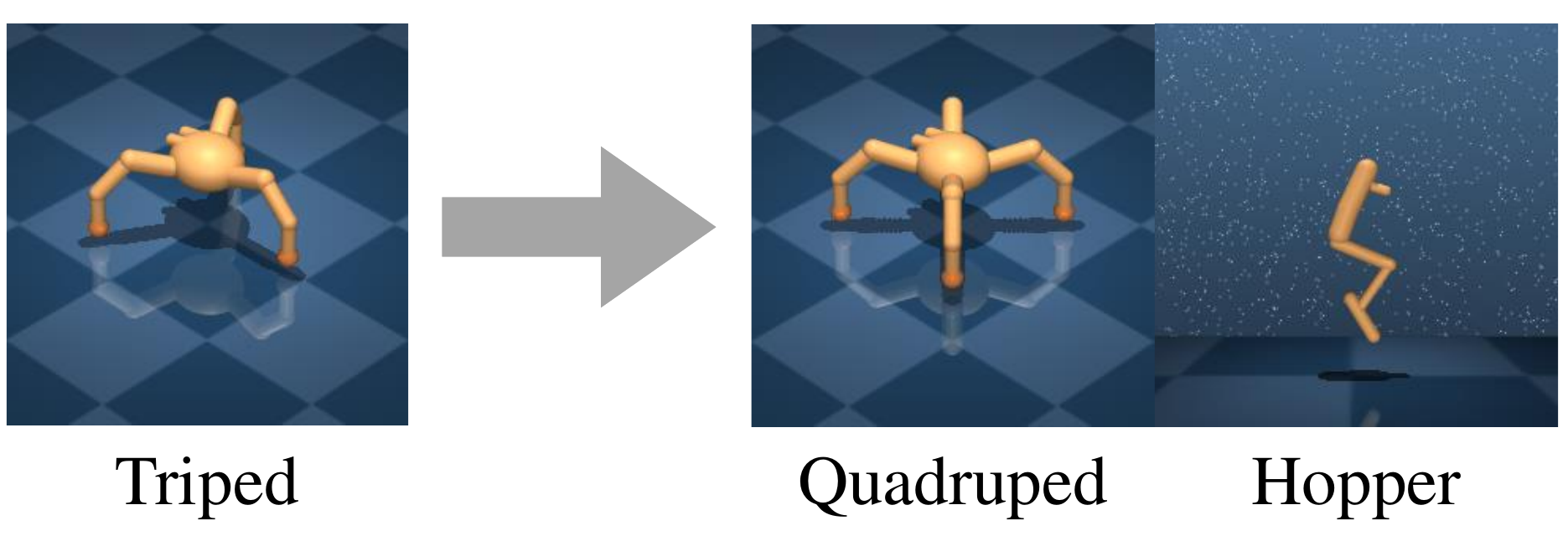}
    \vspace{-0.1in}
    \caption{
    Illustration of additional experimental setup in DeepMind Control Suite experiments. We use videos collected in Triped Walk for pre-training, and then fine-tune the pre-trained model for solving downstream Quadruped and Hopper locomotion tasks.}
    \label{fig:setup_triped}
    \vspace{-0.1in}
\end{figure}

\paragraph{DeepMind Control Suite experiments.} We also consider widely used robotic locomotion tasks from DeepMind Control Suite~\citep{tassa2020dm_control}.
Following the common setup in this benchmark~\citep{hafner2019dream}, the episode length is 1000 steps with the action repeat of 2, and reward ranges from 0 to 1.
For pre-training, we consider two datasets: (i) 1000 videos collected from Triped Walk (see~\cref{fig:setup_triped}) and (ii) manipulation videos from RLBench.
The former one is for evaluating the performance of \ALG on in-domain transfer setup similar to the setup in \citet{stooke2021decoupling}, while the latter one is for investigating whether the pre-trained representations can be transferred to extremely different domains, i.e., out-of-domain transfer.
Specifically, we collect 1000 videos encountered during the training of DreamerV2 agent in Triped Walk and use these videos for pre-training.
For downstream tasks, we fine-tune the model for 1M environment steps.
See~\cref{appendix:experimental_details} for more details.

\paragraph{Hyperparameters.}
For newly introduced hyperparameters, we use $\beta_{z}=1.0$ for pre-training, and $\beta_{z}=0, \beta=1.0$ for fine-tuning.
We use $\tau=5$ for computing the intrinsic bonus. To make the scale of intrinsic bonus be 10\% of extrinsic reward, we normalize the intrinsic reward and use $\lambda = 0.1, 1.0$ for manipulation and locomotion tasks, respectively.
We find that increasing the hidden size of dense layers and the model state dimension from 200 to 1024 improves the performance of both \ALG and DreamerV2.
We use $T = 25, 50$ for manipulation and locomotion tasks, respectively, during pre-training.
Unless otherwise specified, we use the default hyperparameters of DreamerV2.

\subsection{Meta-world Experiments}
\paragraph{RLBench pre-training results.}
\cref{fig:metaworld_performance} shows the learning curves of \ALG pre-trained using the RLBench videos on six robotic manipulation tasks from Meta-world.
We find that \ALG consistently outperforms DreamerV2 in terms of sample-efficiency in all considered tasks. In particular, our framework achieves success rate above 60\% on Lever Pull task while DreamerV2 completely fails to solve the task.
These results show that \ALG can leverage action-free videos for learning useful representations that improve the sample-efficiency of vision-based RL.

\paragraph{Comparison with DrQ-v2.}
We also compare \ALG with a state-of-the-art model-free RL method DrQ-v2~\citep{yarats2021mastering} in~\cref{appendix:metaworld_drqv2}.
We find that \ALG also significantly outperforms DrQ-v2 on most tasks, while DrQ-v2 struggles to achieve strong performance.
While this aligns with the observation of \citet{yarats2021mastering} where DreamerV2 outperformed DrQ-v2 on DeepMind Control Suite, investigating why DrQ-v2 fails is an interesting future direction.

\begin{figure*} [t!] \centering
\subfigure[Comparison with na\"ive fine-tuning]
{
\includegraphics[width=0.315\textwidth]{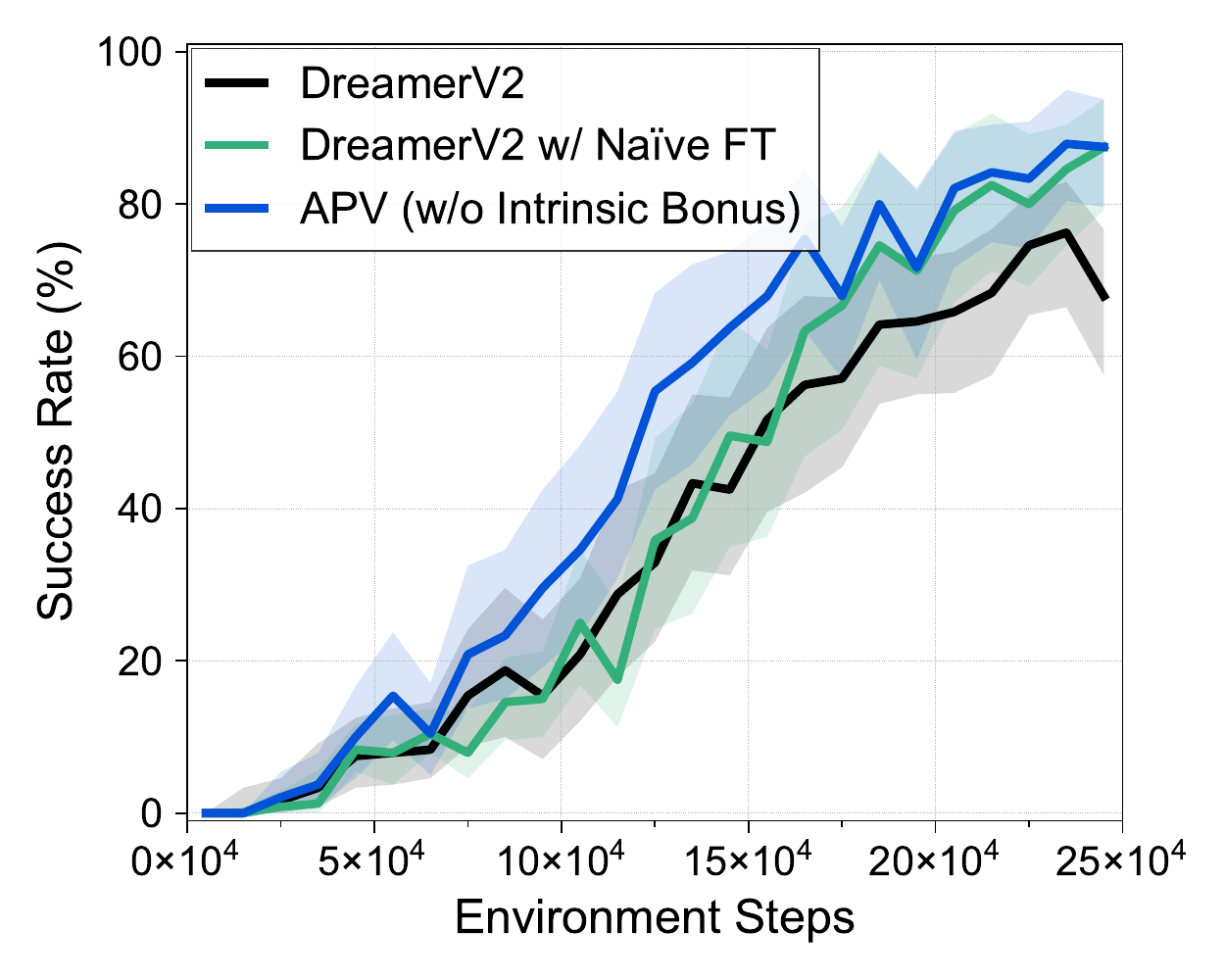}
\label{fig:metaworld_naive_finetuning}}
\subfigure[Effects of pre-training and intrinsic bonus]
{
\includegraphics[width=0.315\textwidth]{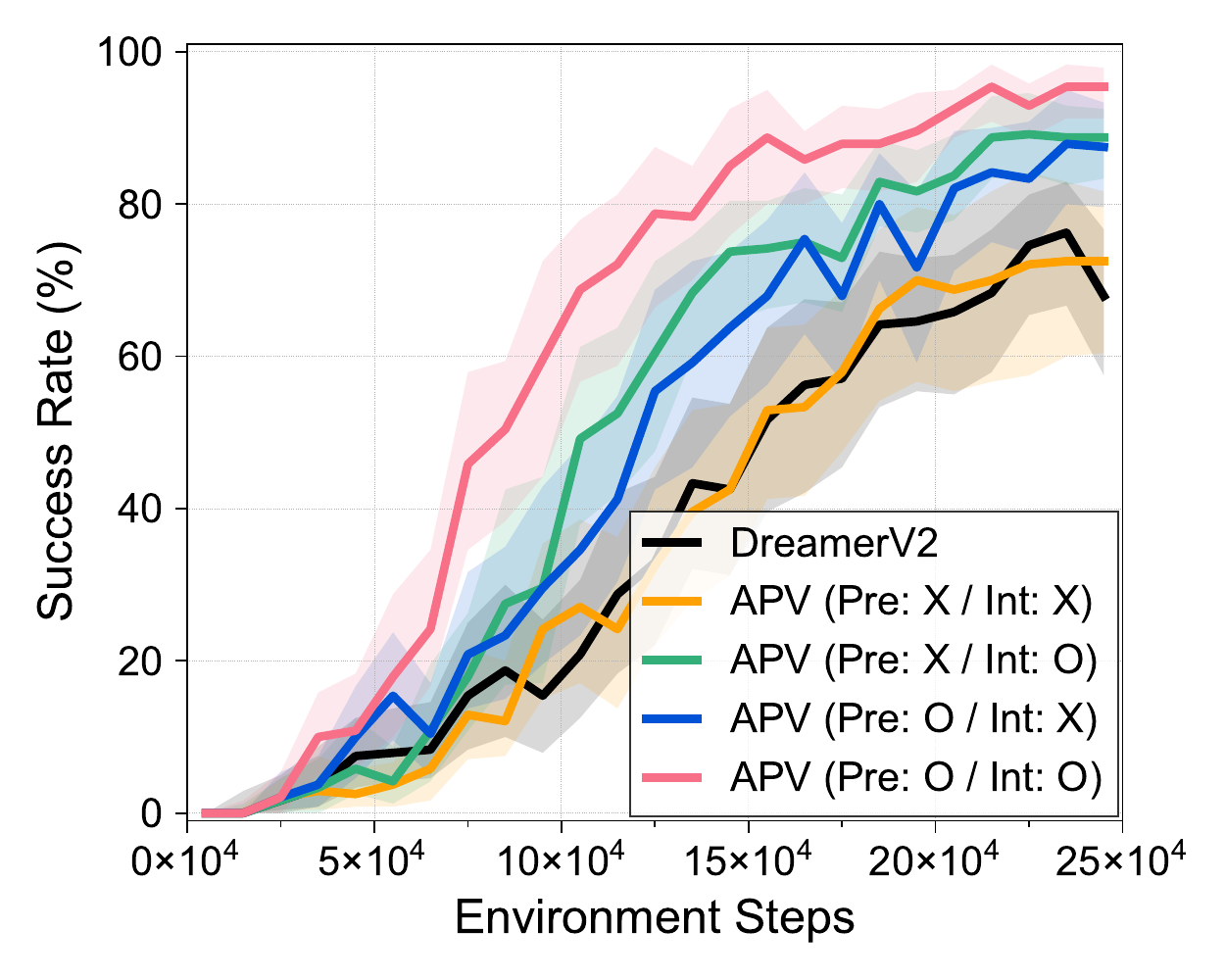}
\label{fig:metaworld_ablation}}
\subfigure[Length of future model states $\tau$]
{
\includegraphics[width=0.315\textwidth]{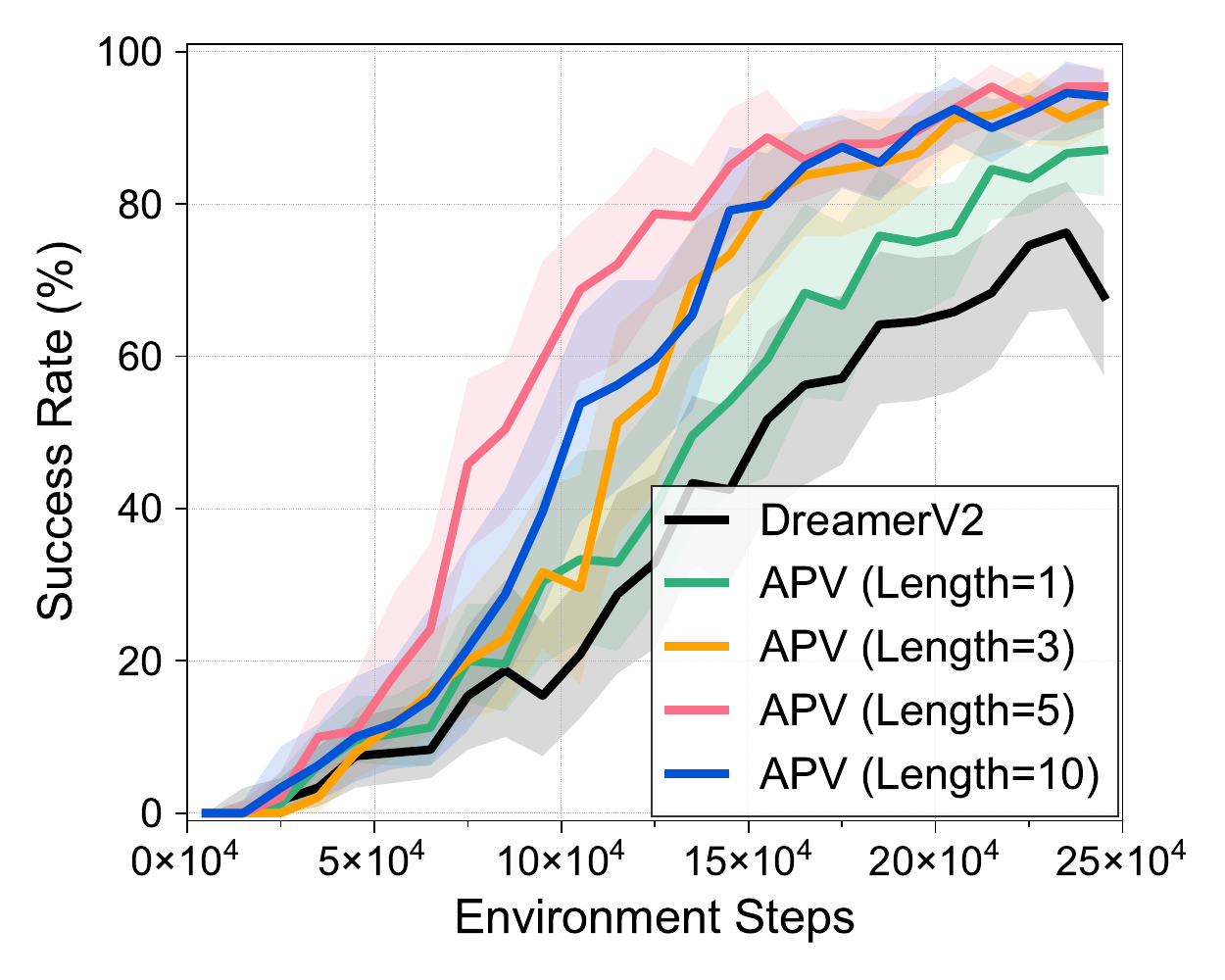}
\label{fig:metaworld_intrinsic}}
\vspace{-0.125in}
\caption{
Learning curves on manipulation tasks from Meta-world as measured on the success rate. We report the interquartile mean and stratified bootstrap confidence interval across total 48 runs over six tasks.
(a) Comparison with a na\"ive fine-tuning scheme that initializes the action-conditional prediction model with the action-free prediction model. (b) Performance of \ALG with or without generative pre-training and intrinsic bonus.
Here, \textit{Pre} denotes generative pre-training, and \textit{Int} denotes intrinsic bonus.
(c) Performance of \ALG with varying the length of future model states $\tau$ used for computing the intrinsic bonus.
}
\vspace{-0.2in}
\label{fig:metaworld_analysis}
\end{figure*}

\begin{figure*} [t!] \centering
\subfigure[Video representations]
{
\includegraphics[width=0.315\textwidth]{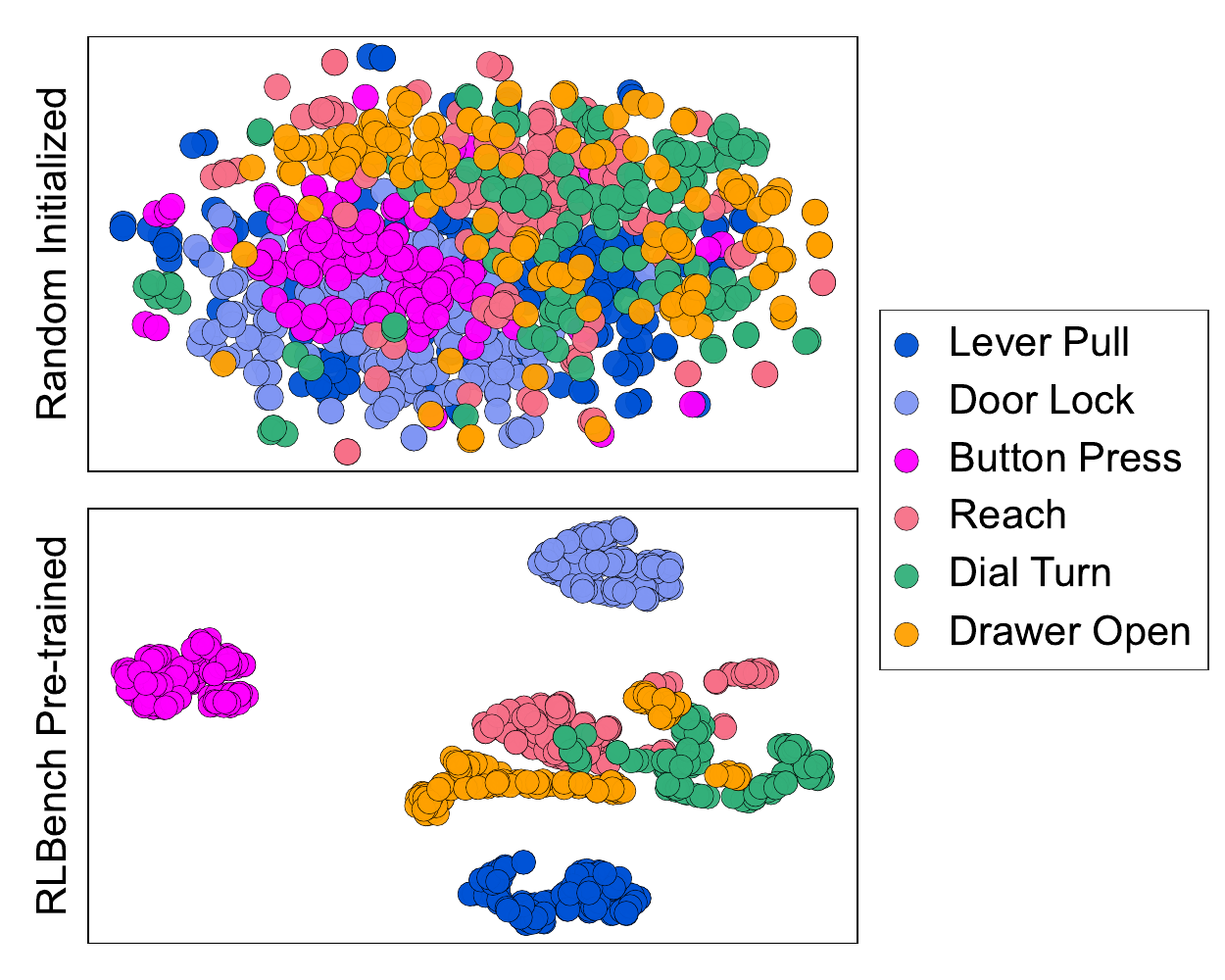}
\label{fig:metaworld_tsne}}
\subfigure[Importance of dynamics information]
{
\includegraphics[width=0.315\textwidth]{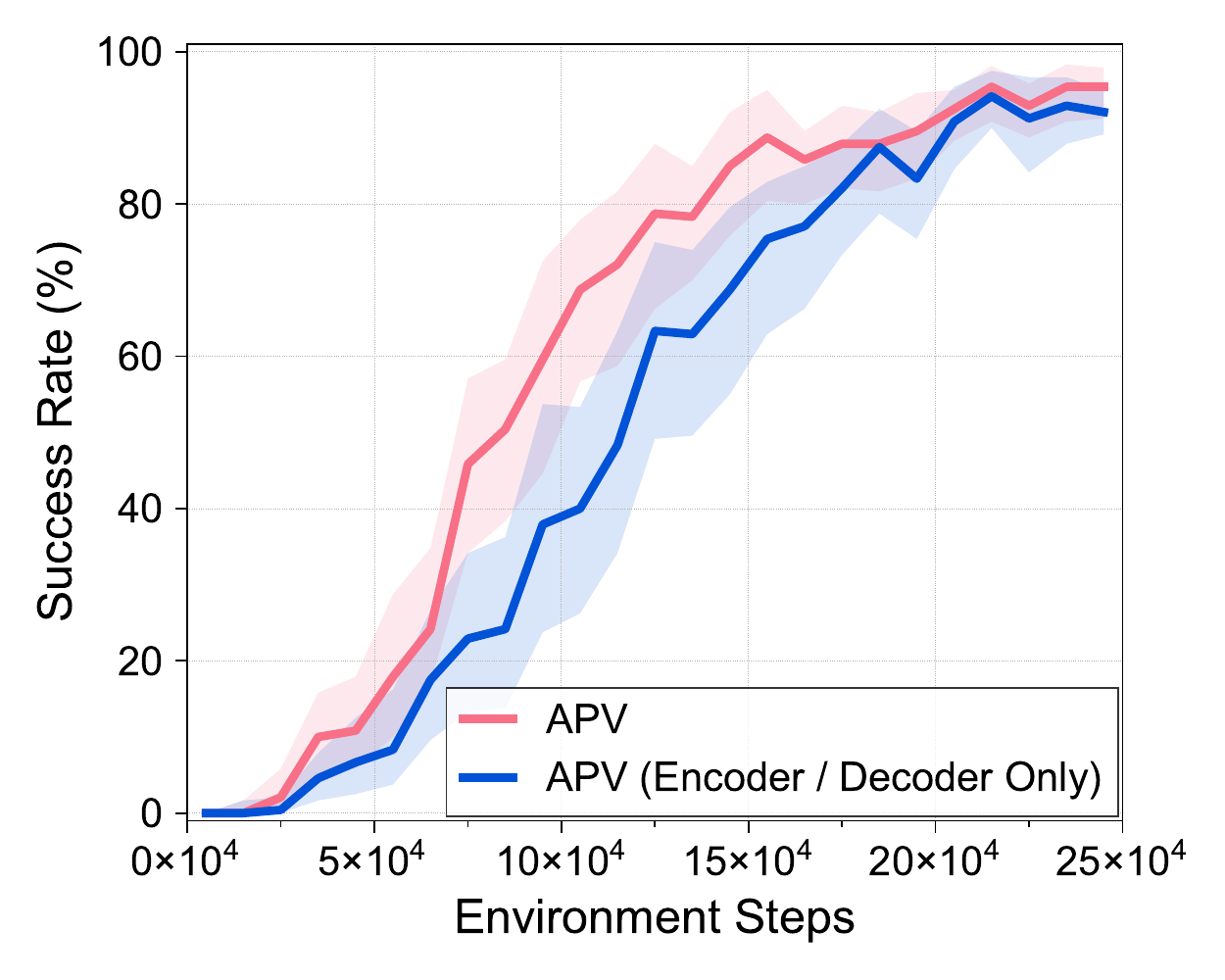}
\label{fig:metaworld_apv_norssm}}
\subfigure[Effects of in-domain videos]
{
\includegraphics[width=0.315\textwidth]{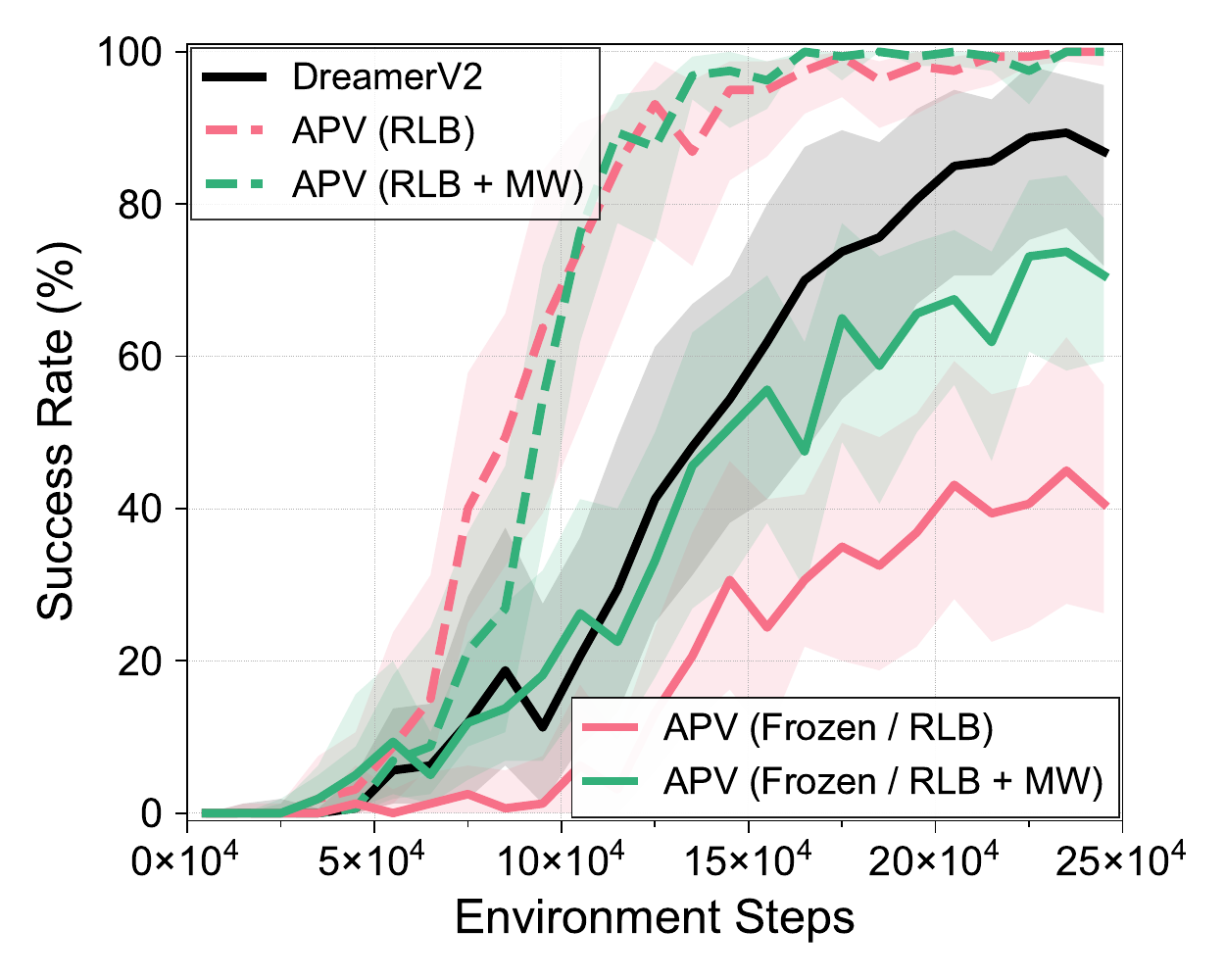}
\label{fig:metaworld_mw_rlb}}
\vspace{-0.125in}
\caption{
(a) t-SNE visualization of average pooled model states from the action-free prediction model.
(b) Learning curves of \ALG on six manipulation tasks when only the parameters of convolutional image encoder and decoder are transferred, i.e., without transferring dynamics information captured in recurrent models.
(c) Learning curves of \ALG on four manipulation tasks when pre-trained on RLBench videos only (RLB), and on both of RLBench videos and additional in-domain Meta-world videos (RLB + MW).
Dotted and bold lines indicate the performance when all parameters are fine-tuned, and the representation model of the action-free model is frozen, respectively.
}
\label{fig:metaworld_analysis2}
\vspace{-0.1in}
\end{figure*}

\paragraph{Comparison with na\"ive fine-tuning.}
To verify the necessity of the proposed architecture for fine-tuning, we compare \ALG to a
na\"ive fine-tuning scheme that initializes the action-conditional latent dynamics model with the pre-trained parameters of the action-free model (see~\cref{appendix:experimental_details} for the details).
For a fair comparison, we do not utilize the intrinsic bonus for \ALG.
\cref{fig:metaworld_naive_finetuning} shows that DreamerV2 with this na\"ive fine-tuning scheme (DreamerV2 w/ Na\"ive FT) does not provide large gains over DreamerV2, which implies that na\"ive fine-tuning quickly loses pre-trained representations.
By contrast, we find that \ALG without intrinsic bonus consistently outperforms DreamerV2 by achieving $>10\%$ higher success rate from the beginning of the fine-tuning, even though the same pre-trained model is used for fine-tuning.
This shows the proposed architecture is crucial for effective fine-tuning.

\paragraph{Ablation study.}
To evaluate the contribution of the proposed techniques in \ALG, we report the performance of our framework with or without generative pre-training and intrinsic bonus in~\cref{fig:metaworld_ablation}.
First, we observe that \ALG (Pre: X / Int: X), whose difference to DreamerV2 is the usage of stacked latent prediction model, achieves similar performance to DreamerV2.
This implies that our performance gain is not from the architecture itself, but the way we utilize it for fine-tuning is important.
We also find that our intrinsic bonus can improve the performance with or without pre-training, and generative pre-training can also improve the performance with or without intrinsic bonus.
Importantly, one can see that the best performance is achieved when both components are combined.
This implies that our proposed techniques synergistically contribute to the performance improvement.

\paragraph{Effects of video-based intrinsic bonus.}
We investigate the effect of considering multiple model states of length $\tau$ in \cref{eqn:intrinsic_bonus} instead of a single model state.
\cref{fig:metaworld_intrinsic} shows that \ALG with $\tau=5$ achieves better performance than $\tau\in\{1,3\}$.
We think this is because considering a sequence of observations, i.e., videos, enables us to utilize contextual information for encouraging agents to perform diverse behaviors.
But we also find that \ALG with $\tau=10$ performs worse than $\tau=5$, which might be due to the complexity from average pooling over longer videos.

\paragraph{Qualitative analysis.}
We visually investigate why the pre-trained representations can be useful for unseen Meta-world tasks.
Specifically, we sample video clips of length 25 from the 10 videos of six manipulation task, and visualize the averaged model states from the sampled videos using t-SNE \citep{van2008visualizing} in \cref{fig:metaworld_tsne}, where colors indicate the tasks.
We find that the pre-trained representations from each task are clustered, while randomly initialized ones are entangled.
This shows that the pre-trained representations capture information about the tasks without access to Meta-world videos during pre-training.

\begin{figure*} [t!] \centering
\subfigure[State regression]
{
\includegraphics[width=0.315\textwidth]{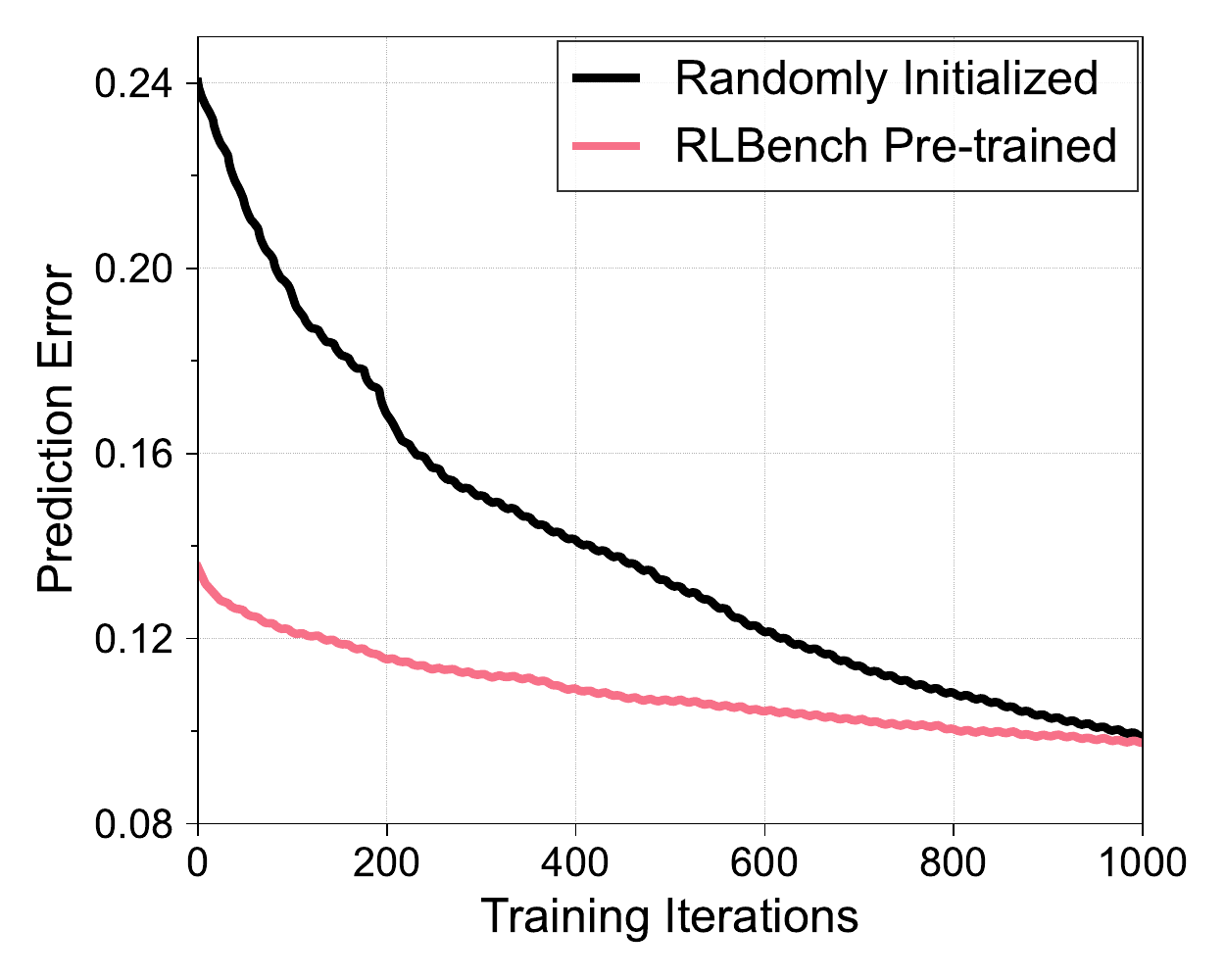}
\label{fig:state_regression}}
\subfigure[Reward regression]
{
\includegraphics[width=0.315\textwidth]{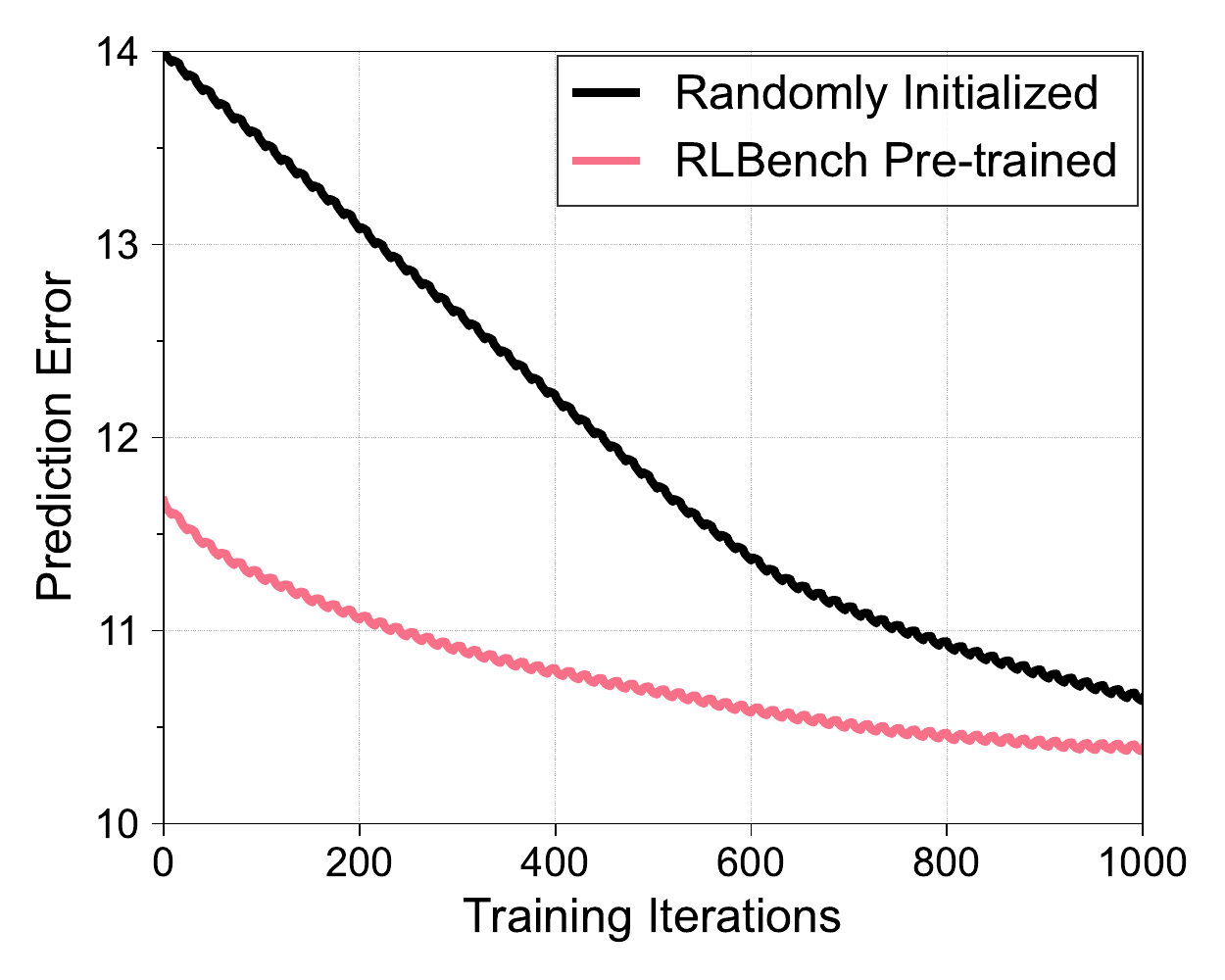}
\label{fig:reward_regression}}
\subfigure[Effects of pre-training datasets]
{
\includegraphics[width=0.315\textwidth]{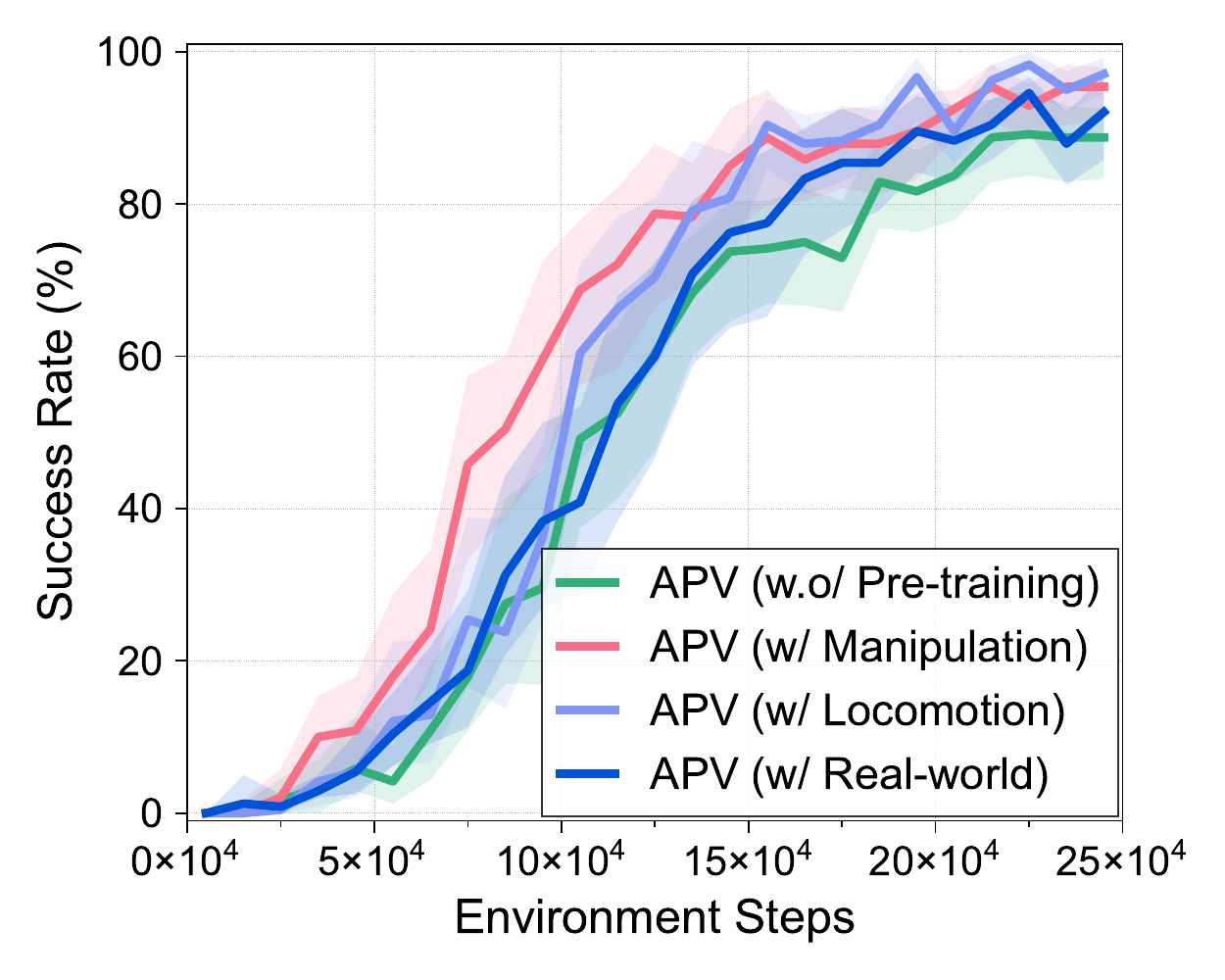}
\label{fig:metaworld_dataset}}
\vspace{-0.125in}
\caption{
We report the prediction error on held-out test sets obtained while training a regression model to predict (a) proprioceptive states and (b) rewards. We observe that RLBench pre-trained model achieves a small prediction error throughout training from the beginning. (c) Learning curves on manipulation tasks from Meta-world when pre-trained with manipulation, locomotion, and real-world videos. We report the interquartile mean and stratified bootstrap confidence interval over 48 runs over six tasks.
}
\label{fig:analysis3}
\vspace{-0.1in}
\end{figure*}

\begin{figure*} [t!] \centering
\includegraphics[width=0.99\textwidth]{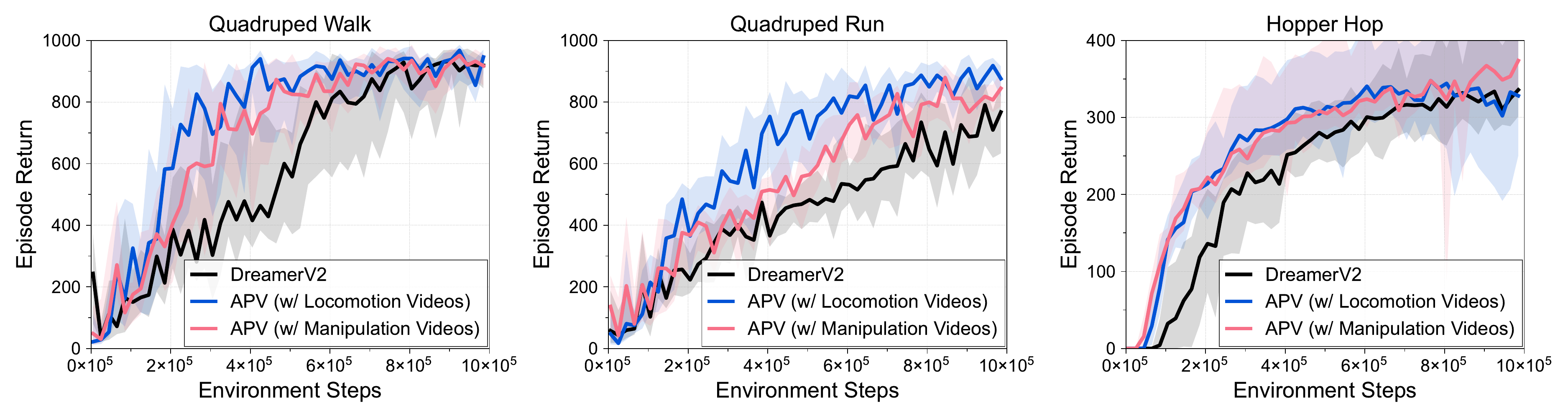}
\vspace{-0.15in}
\caption{
Learning curves on locomotion tasks from DeepMind Control Suite as measured on the episode return.
Interestingly, we find that \ALG pre-trained on manipulation videos from RLBench consistently outperforms DreamerV2.
We also observe that utilizing in-domain videos from Triped Walk leads to further improved performance.
The solid line and shaded regions represent the interquartile mean and bootstrap confidence intervals, respectively, across eight runs.}
\label{fig:dmc}
\vspace{-0.125in}
\end{figure*}

\paragraph{Importance of dynamics information.}
To investigate whether the performance gain comes from utilizing the dynamics information captured in pre-trained representations or visual information captured in the image encoder and decoder,
we report the performance of \ALG when only the pre-trained parameters of the convolutional image encoder and decoder are transferred.
\cref{fig:metaworld_apv_norssm} shows that \ALG (Encoder / Decoder Only) performs worse than \ALG, which demonstrates that utilizing the dynamics information learned from diverse videos is crucial for performance.

\paragraph{Pre-training with additional in-domain videos.}
We also consider an experimental setup where we have an access to additional videos with similar visuals collected on the target domain.
Specifically, we collect 10 videos on each task from ML-10 training tasks in Meta-world, i.e., total of 100 videos, and utilize these videos for pre-training in conjunction with RLBench videos.
Then, we fine-tune the pre-trained model for solving 4 manipulation tasks that were not seen during pre-training.
To evaluate how additional videos affect representation learning, we report the performance of \ALG when the representation model of the action-free prediction model is frozen, and use it as a proxy for evaluating the quality of representations.

\cref{fig:metaworld_mw_rlb} shows that pre-training with additional Meta-world videos (\ALG (RLB + MW)) achieves almost similar performance to pre-training with only RLBench videos (\ALG (RLB)).
This shows that pre-trained representations with RLBench videos already learns useful representations so that they can be quickly fine-tuned for solving Meta-world tasks.
However, we observe that pre-training with in-domain videos significantly improves the performance when the pre-trained representation model is frozen. This implies that additional in-domain videos help for addressing the domain gap between pre-training and fine-tuning.

\paragraph{State and reward regression analysis.} Our hypothesis for how pre-training from videos is useful for RL is that pre-trained representations can capture useful information (e.g., rewards and proprioceptive states) for solving RL tasks.
To test this hypothesis, we train a regression model that predicts proprioceptive states and rewards on the pre-collected Triped dataset. \cref{fig:state_regression} and~\cref{fig:reward_regression} show the regression performances with and without RLBench pre-trained representations. We find that the RLBench pre-trained model not only has a small prediction error at the beginning, but also quickly converges compared to the model trained from scratch.
This shows that pre-training helps the agent to understand environment.

\paragraph{Effects of pre-training datasets}
To investigate the effect of pre-training domains, we evaluate the performance of \ALG on Meta-world when pre-trained with different datasets.
To this end, we consider pre-training on manipulation videos from RLBench (i.e., \ALG w/ Manipulation Videos), locomotion videos from Triped dataset (i.e., \ALG w/ Locomotion Videos), and real-world natural videos of people performing diverse behaviors (i.e., \ALG w/ Real-world).
Specifically, for real-world videos, we utilize Something-Something-V2 dataset \citep{goyal2017something}, which contains 159K videos of people performing actions.
Because we find that our video prediction model suffers from severe underfitting to real-world video datasets (see~\cref{appendix:video_prediction_something} for the examples of blurry predicted future frames),
we use subsampled 1.5K videos for pre-training.
Moreover, it would be interesting to study whether pre-training on complex robotics video (e.g., RoboNet~\citep{dasari2019robonet}) can improve performance on more complex robotics tasks (e.g., RLBench).

In \cref{fig:metaworld_dataset}, we observe that pre-training with videos from more similar domains can be helpful.
For instance, \ALG w/ Manipulation outperforms APV w/ Locomotion, which shows that pre-training with manipulation videos can be more effective for manipulation tasks.
However, APV w/ Real-world struggles to outperform the baseline without pre-training even though there is no underfitting issue,
which might be due to the small number of training data and domain gap between egocentric real-world videos and third-person robotic videos.
It would be an interesting direction to investigate how different aspects of pre-training datasets, e.g., size, domain, and point of view.

\subsection{DeepMind Control Suite Experiments}
\cref{fig:dmc} shows the learning curves of \ALG and DreamerV2 on locomotion tasks.
Interesingly, we find that \ALG pre-trained on manipulation videos from RLBench (pink curves) consistently achieves better performance than DreamerV2 (black curves).
This demonstrates that representations pre-trained using manipulation videos, which have notably different visuals and objectives, effectively capture dynamics information useful for quickly learning the dynamics of locomotion environments.
Also, by utilizing the videos from a similar domain, i.e., Triped environment, the performance of \ALG is further improved.
We provide additional experimental results that evaluate the contribution of the proposed techniques on DeepMind Control Suite tasks in \cref{appendix:dmc_ablation}.

\section{Discussion}
In this work, we introduce a vision-based RL framework that learns representations useful for understanding the dynamics of downstream domains via action-free pre-training on videos, and utilizes the pre-trained representations for fine-tuning.
Our experimental results demonstrate that \ALG can improve both sample-efficiency and final performances of vision-based RL on various manipulation and locomotion tasks, by effectively transferring the pre-trained representations from unseen domains.
However, one limitation of our work is that pre-training is conducted only on simulated robotic videos, which is because of the underfitting issue reported in our analysis.
Given that, one interesting direction would be to scale up the architecture or utilize recently developed video prediction architectures~\citep{babaeizadeh2021fitvid,yan2021videogpt}, and investigate how the prediction quality affects the performance.
Another interesting direction is to incorporate generalization approaches in RL~\citep{tobin2017domain,higgins2017darla,laskin2020reinforcement} to deal with the difference between pre-training and fine-tuning domains.
Moreover, while our work focuses on representation learning via generative pre-training, another interesting future direction would be to investigate the performance of representation learning schemes such as masked prediction~\citep{he2021masked,xiao2022masked,yu2022mask}, latent reconstruction~\citep{yu2021playvirtual,schwarzer2020data}, and contrastive learning~\citep{oord2018representation}.
It would also be interesting to utilize our pre-trained representations for tasks that require more long-term reasoning, such as reward learning from preferences~\citep{park2022surf} or videos~\citep{chen2021learning}.
By presenting a generic framework that can leverage videos with diverse visuals and embodiments for pre-training, we hope this work would facilitate future research on unsupervised pre-training for RL.

\section*{Acknowledgements}
We would like to thank Danijar Hafner, Junsu Kim, Fangchen Liu, Jongjin Park, Jihoon Tack, Wilson Yan, and Sihyun Yu for helpful discussions.
We also thank Cirrascale Cloud Services\footnote{\url{https://cirrascale.com}} for providing compute resources.
This work was partially supported by Hong Kong Centre for Logistics Robotics, Center for Human Compatible AI (CHAI), Institute of Information \& Communications Technology Planning \& Evaluation (IITP) grant funded by the Korea government (MSIT) (No.2019-0-00075, Artificial Intelligence Graduate School Program (KAIST)), and Institute of Information \& communications Technology Planning \& Evaluation (IITP) grant funded by the Korea government(MSIT) (No.2022-0-00166, Self-directed AI Agents with Problem-solving Capability).

\bibliography{main}
\bibliographystyle{icml2022}

\newpage
\appendix
\onecolumn
\section{Behavior Learning }
\label{appendix:behavior_learning}
For behavior learning, we utilize actor-critic learning scheme of \citet{hafner2020mastering} where the agent maximizes the values of imagined future states by propagating the analytic gradients back through the world model separately learned with \cref{eq:stacked_world_model}.
Specifically, given a stochastic actor and a deterministic critic as below:
\begin{gather}
\begin{aligned}
&\text{Actor:} &&\hat{a}_{t} \sim p_{\psi}(\hat{a}_{t}\,|\,\hat{s}_{t}) \\
&\text{Critic:} &&v_{\xi}(\hat{s}_{t}) \approx \mathbb{E}_{p_{\theta},p_{\phi}}\left[\textstyle\sum_{i \leq t}\gamma^{i - t} \hat{r}_{i}\right],
\label{eq:actor_critic}
\end{aligned}
\end{gather}
a sequence of H future states $\hat{s}_{1:H}$ is recursively predicted conditioned on initial states $\hat{s}_{0}$, which are model states encountered during world model training, using the stochastic actor and the transition predictor of world model.
Then a deterministic critic is learned to regress the $\lambda$-target~\citep{schulman2015high,sutton2018reinforcement} as follows:
\begin{align}
    \mathcal{L}(\xi)\doteq\mathbb{E}_{p_{\theta},p_{\psi}}\left[\sum^{H-1}_{t=1} \frac{1}{2} \left(v_{\xi}(\hat{s}_{t}) - \text{sg}(V_{t}^{\lambda})\right)^{2}\right],
    \label{eq:critic_loss}
\end{align}
where $\text{sg}$ is a stop gradient function, and $\lambda$-return $V_{t}^{\lambda}$ is defined using the future states as follows:
\begin{align}
    V_{t}^{\lambda}\doteq \hat{r}_{t} + \gamma
    \begin{cases}
      (1 - \lambda)v_{\xi}(\hat{s}_{t+1})+\lambda V_{t+1}^{\lambda} & \text{if}\ t<H \\
      v_{\xi}(\hat{s}_{H}) & \text{if}\ t=H
    \end{cases}
    \label{eq:lambda_return}
\end{align}
Then, the actor is trained to maximize the $\lambda$-return in \cref{eq:critic_loss} by leveraging the straight-through estimator \citep{bengio2013estimating} for backpropagating the value gradients through the discrete world model as follows:
\begin{align}
    \mathcal{L}(\psi)\doteq \mathbb{E}_{p_{\theta},p_{\psi}} \left[-V_{t}^{\lambda} - \eta\,\text{H}\left[a_{t}|\hat{s}_{t}\right] \right],
    \label{eq:actor_loss}
\end{align}
where the entropy of actor $\,\text{H}\left[a_{t}|\hat{s}_{t}\right]$ is maximized to encourage exploration, and $\eta$ is a hyperparameter that adjusts the strength of entropy regularization.

\section{Formulation with Recurrent State-Space Model}
\label{appendix:formulation_rssm}
While we follow the formulation of \citet{hafner2019dream} for our main draft, we additionally provide a detailed formulation of the action-free latent video prediction model and stacked latent prediction model implemented with a recurrent state-space model (RSSM; \citealt{hafner2019learning}) that consists of deterministic and stochastic components, for better understanding of our implementation.

\subsection{Action-free Latent Video Prediction Model}
\label{appendix:formulation_rssm_action_free}
The main component of our model is an action-free RSSM that consists of (i) a recurrent model that computes deterministic states $h^{\tt{AF}}_{t}$ conditioned on previous states for each time step $t$, (ii) a representation model that computes posterior stochastic states $z^{\tt{AF}}_{t}$ conditioned on determinsitic states $h^{\tt{AF}}_{t}$ and image observations $o_{t}$, (iii) a transition predictor that computes prior stochastic states $\hat{z}^{\tt{AF}}_{t}$ without access to image observations.
We define the model state as the concatenation of $h^{\tt{AF}}_{t}$ and $z^{\tt{AF}}_{t}$, which is used as an input to the decoder.
The model can be summarized as:
\begin{gather}
\begin{aligned}
&\text{Recurrent model:} &&h^{\tt{AF}}_t=f_\phi(h^{\tt{AF}}_{t-1},z^{\tt{AF}}_{t-1}) \\
&\text{Representation model:} &&z^{\tt{AF}}_t\sim q_\phi(z^{\tt{AF}}_{t} \,|\,h^{\tt{AF}}_{t},o_{t}) \\
&\text{Transition predictor:} &&\hat{z}^{\tt{AF}}_t\sim p_\phi(\hat{z}^{\tt{AF}}_{t} \,|\,h^{\tt{AF}}_{t}) \\
&\text{Image decoder:} &&\hat{o}_t\sim p_\phi(\hat{o}_{t} \,|\,h^{\tt{AF}}_{t}, z^{\tt{AF}}_{t})
\label{eq:appendix_action_free}
\end{aligned}
\end{gather}
At inference time, the recurrent model and the transition predictor (i.e., learned prior) are used for predicting future model states, from which the image decoder reconstruct future frames.
All the components of the model parameterized by $\phi$ are jointly optimized by minimizing the following loss:
\begin{align}
    \mathcal{L}(\phi) &\doteq \;\mathbb{E}_{q_{\phi}\left(z^{\tt{AF}}_{1:T}\,|\,o_{1:T}\right)}\Big[
    \textstyle\sum_{t=1}^{T} \Big(
    \describe{-\ln p_{\phi}(o_{t}\,|\,h^{\tt{AF}}_{t},z^{\tt{AF}}_{t})}{image log loss} \describe{+\beta^{\tt{AF}}\,\text{KL}\left[q_{\phi}(z^{\tt{AF}}_{t}\,|\,h^{\tt{AF}}_{t},o_{t}) \,||\,q_{\phi}(\hat{z}^{\tt{AF}}_{t}\,|\,h^{\tt{AF}}_{t}) \right]}{action-free RSSM KL loss}
    \Big)\Big], \label{eq:appendix_action_free_objective}
\end{align}
which is a negative variational lower bound (ELBO; \citealt{kingma2013auto}) objective where $\beta^{\tt{AF}}$ is a scale hyperparameter and $T$ is the length of training sequences in a minibatch.

\subsection{Stacked Latent Prediction Model}
\label{appendix:formulation_rssm_stacked}
Our stacked latent prediction model consists of the action-free RSSM followed by the action-conditional RSSM as below:
\begin{align}
&\text{\textbf{Action-free RSSM}} \nonumber\\[-0.04in]
\raisebox{1.95ex}{\llap{\blap{\ensuremath{ \hspace{0.1ex} \begin{cases} \hphantom{A} \\ \hphantom{A} \\ \hphantom{A} \end{cases} \hspace*{-4ex}
}}}}
&\text{Recurrent model:} &&h^{\tt{AF}}_t=f_\phi(h^{\tt{AF}}_{t-1},z^{\tt{AF}}_{t-1}) \nonumber\\
&\text{Representation model:} &&z^{\tt{AF}}_t\sim q_\phi(z^{\tt{AF}}_{t} \,|\,h^{\tt{AF}}_{t},o_{t}) \nonumber\\
&\text{Transition predictor:} &&\hat{z}^{\tt{AF}}_t\sim p_\phi(\hat{z}^{\tt{AF}}_{t} \,|\,h^{\tt{AF}}_{t}) \nonumber\\[0.05in]
&\text{\textbf{Action-conditional RSSM}} \nonumber\\[-0.05in]
\raisebox{1.95ex}{\llap{\blap{\ensuremath{ \hspace{0.1ex} \begin{cases} \hphantom{A} \\ \hphantom{A} \\ \hphantom{A} \end{cases} \hspace*{-4ex}
}}}}
&\text{Recurrent model:} &&h^{\tt{AC}}_t=f_\theta(h^{\tt{AC}}_{t-1},z^{\tt{AC}}_{t-1},a_{t-1}) \nonumber\\
&\text{Representation model:} &&z^{\tt{AC}}_t\sim q_\theta(z^{\tt{AC}}_{t} \,|\,h^{\tt{AC}}_{t},a_{t-1},h^{\tt{AF}}_{t},z^{\tt{AF}}_{t}) \nonumber\\
&\text{Transition predictor:} &&\hat{z}^{\tt{AC}}_t\sim p_\theta(\hat{z}^{\tt{AC}}_{t} \,|\,h^{\tt{AC}}_{t}) \nonumber\\[0.05in]
&\text{Image decoder:} &&\hat{o}_t\sim p_\phi(\hat{o}_{t} \,|\,h^{\tt{AC}}_{t}, z^{\tt{AC}}_{t}) \nonumber\\
&\text{Reward predictor:} &&\hat{r}_t\sim p_\theta(\hat{r}_{t} \,|\,h^{\tt{AC}}_{t}, z^{\tt{AC}}_{t})
\label{eq:appendix_stacked_world_model}
\end{align}
All components parameterized by $\phi$ and $\theta$ are optimized jointly by minimizing the following:
\begin{align}
    &\mathcal{L}(\phi,\theta) \doteq  \mathbb{E}_{q_{\theta}\left(z^{\tt{AC}}_{1:T}\,|\,a_{1:T},h^{\tt{AF}}_{1:T},z^{\tt{AF}}_{1:T}\right),q_{\phi}\left(z^{\tt{AF}}_{1:T}\,|\,o_{1:T}\right)}\Big[\textstyle\sum_{t=1}^{T} \Big(
    \describe{-\ln p_{\theta}(o_{t}\,|\,h^{\tt{AC}}_{t},z^{\tt{AC}}_{t})}{image log loss}
    \describe{-\ln p_{\theta}(r_{t}\,|\,h^{\tt{AC}}_{t},z^{\tt{AC}}_{t})}{reward log loss} \nonumber \\
    & \describe{+ \beta^{\tt{AF}}\,\text{KL}\left[ q_{\phi}(z^{\tt{AF}}_{t}\,|\,h^{\tt{AF}}_{t},o_{t}) \,||\,q_{\phi}(\hat{z}^{\tt{AF}}_{t}\,|\,h^{\tt{AF}}_{t}) \right]}{action-free RSSM KL loss} \describe{+ \beta^{\tt{AC}}\,\text{KL}\left[ q_{\theta}(z^{\tt{AC}}_{t}\,|\,h^{\tt{AC}}_{t},h^{\tt{AF}}_{t},z^{\tt{AF}}_{t}) \,||\, q_{\theta}(\hat{z}^{\tt{AC}}_{t}\,|\,h^{\tt{AC}}_{t}) \right]}{action-conditional RSSM KL loss}
    \Big)\Big], \label{eq:appendix_stacked_world_model_objective}
\end{align}
where $\beta^{\tt{AC}}$ is a scale hyperparameter.

\section{Extended Related Work}
\label{appendix:extended_related_work}
\paragraph{Video prediction.}
A line of works close to our work is video prediction methods, that aims to predict the future frames conditioned on images~\citep{michalski2014modeling,ranzato2014video,srivastava2015unsupervised,vondrick2016generating,lotter2016deep}, texts~\citep{wu2021godiva}, and actions~\citep{oh2015action,finn2016unsupervised}, which would be useful for various applications, e.g., control with world models~\citep{hafner2019learning,kaiser2019model,rybkin2021model}, and simulator development~\citep{kim2020learning,kim2021drivegan}.
Recent successful approaches include generative adversarial networks (GANs;~\citealt{goodfellow2014generative}) known to generative a sequence of frames by introducing adversarial discriminators that makes prediction based on temporal information~\citep{aigner2018futuregan,jang2018video,kwon2019predicting,clark2019adversarial,luc2020transformation}, and autoregressive video prediction models~\citep{babaeizadeh2017stochastic,kalchbrenner2017video,reed2017parallel, denton2018stochastic,lee2018stochastic,villegas2019high,weissenborn2019scaling,wu2021godiva,babaeizadeh2021fitvid,yan2021videogpt,seo2022autoregressive,yan2022patch}.
Our work builds on prior latent video prediction methods that utilize a state-space model to the domain of video prediction~\citep{hafner2019learning,franceschi2020stochastic}, which enables us to predict future states without conditioning on predicted frames in an efficient manner.
It would be an interesting direction to scale up current architecture or develop a new high-fidelity latent prediction model and utilize it for pre-training for RL.

\paragraph{Exploration in RL.}
Exploration in RL has been studied for encouraging the agents to visit more diverse states by maximizing the entropy of the action space~\citep{ziebart2010modeling,haarnoja2018soft} and the various form of intrinsic bonus, including prediction errors~\citep{houthooft2016vime,pathak2017curiosity,burda2018exploration,sekar2020planning}, count-based state novelty~\cite{bellemare2016unifying, tang2017exploration, ostrovski2017count}, and state entropy~\citep{hazan2019provably,lee2019efficient,liu2021behavior,seo2021state,yarats2021reinforcement}.
Our framework differs in that we explicitly consider a sequence of future model states for computing the intrinsic bonus, while previous works only consider a single state. 
In this work, we show that this encourages the agents to explore the environments in a more long-term manner, thus perform more diverse behaviors.

\clearpage

\section{Difference to DreamerV2}
\label{appendix:dreamerv2_difference}
\paragraph{Architecture.}
The main difference lies in the architecture of world models used for imagining future model states.
\ALG utilizes a stacked latent prediction model where the action-free prediction model first processes $o_{t}$ into representations $z_{t}$ and then the action-conditional prediction model processes them into $s_{t}$ conditioned on additional actions.
In contrast, DreamerV2 utilizes a RSSM where the action-conditional prediction model directly processes $o_{t}$ into model states $s_{t}$ conditioned on actions.
Namely, in terms of architecture, \ALG can be seen as a DreamerV2 that takes representations from the action-free prediction model as inputs instead of raw observations.

\paragraph{Behavior learning.} The behavior learning scheme of \ALG is same as DreamerV2 except that \ALG learns a reward predictor to predict the sum of both extrinsic reward and intrinsic reward. In terms of behavior learning, DreamerV2 can be seen as a specialized case of \ALG with a scale hyperparameter $\lambda = 0$.

\section{Experimental Details}
\label{appendix:experimental_details}
\paragraph{Implementation.}
We build our framework on top of the official implementation of DreamerV2\footnote{\url{https://github.com/danijar/dreamerv2}}, which is based on TensorFlow~\citep{abadi2016tensorflow}.
We use a single Nvidia RTX3090 GPU and 10 CPU cores for each training run. 
The training time required for pre-training of \ALG is 24 hours, and fine-tuning of \ALG requires 4.75 hours for Meta-world experiments, and 6.25 hours for DeepMind Control Suite experiments, when used with XLA optimization.
This takes longer than training of vanilla DreamerV2, which requires 3.5 hours for Meta-world experiments, and 4.25 hours for DeepMind Control Suite experiments.

\paragraph{Dataset details.}
For data collection in Meta-world environment, we use the $\texttt{corner}$ viewpoint for rendering. We use scripted policy availble in the official implementation\footnote{\url{https://github.com/rlworkgroup/metaworld/tree/master/metaworld}}.
For RLBench environment, we use 5 camera viewpoints consisting of $\texttt{front\_rgb}$, $\texttt{left\_shoulder\_rgb}$, $\texttt{overhead\_rgb}$, $\texttt{right\_shoulder\_rgb}$, and $\texttt{wrist\_rgb}$.
We also utilize the scripted policy available in the official implementation\footnote{\url{https://github.com/stepjam/RLBench}}.
For data collection in Triped Walk, we build the tasks by modifying the Quadruped Walk available in DeepMind Control Suite.

\paragraph{Model details.} 
For DeepMind Control Suite experiments, we find that utilizing the concatenation of the model state $z_{t}$ and the representation from the image encoder as inputs to the action-conditional prediction model improves the performance, but in Meta-world experiments, we observe that there is no difference to only utilizing $z_{t}$. Therefore, we only use the concatenation as inputs for DeepMind Control Suite experiments.
For Meta-world experiments, during fine-tuning, we find that not updating the pre-trained video prediction model at the initial phase of DreamerV2 (which is called \textit{pretrain} in the original source code) leads to slightly better performance, so we use this for all Meta-world experiments.
We also find that increasing the hidden size of linear layers and the dimension of deterministic states in RSSMs from 200 to 1024 improves the performance of both our framework and DreamerV2. In all experiments, we use increased model size for all experiments.
Specifically, in our experiments, DreamerV2 has 31M parameters and \ALG has 45M parameters, while DreamerV2 with default hyperparameters has 15M parameters (see~\cref{fig:metaworld_ablation} for the experimental results demonstrating that the performance gain from \ALG is not from the increased number of parameters).
Unless otherwise specified, we follow the architectures and hyperparameters used in \citet{hafner2020mastering}.

\paragraph{Video-based intrinsic bonus details.}
We use $k=16$ for all experiments for computing intrinsic bonus by measuring the distance to $k$-NN representation.
For computing the intrinsic bonus, we construct a queue of size 4096 that contains the recent representations from the action-free model, and use the representations in queue for finding $k$-NN state.
Our preliminary internal results show that this technique of introducing the queue is not critical to the performance, but we leave the technique here, as it is used in our reported experimental results.

\paragraph{Na\"ive fine-tuning details.}
We implement the na\"ive fine-tuning scheme in \cref{fig:metaworld_naive_finetuning} by (i) zero-masking the action inputs during pre-training, and (ii) removing the masks at fine-tuning phase and re-initializing the parameters whose inputs are actions.
We also applied gradient clipping and warm up scheme to both this baseline and DreamerV2, and find that these techniques does not resolve the difficulty of dealing with additional action inputs.

\newpage

\section{Meta-world Experiments with DrQ-v2}
\label{appendix:metaworld_drqv2}
\begin{figure*} [h] \centering
\includegraphics[width=0.99\textwidth]{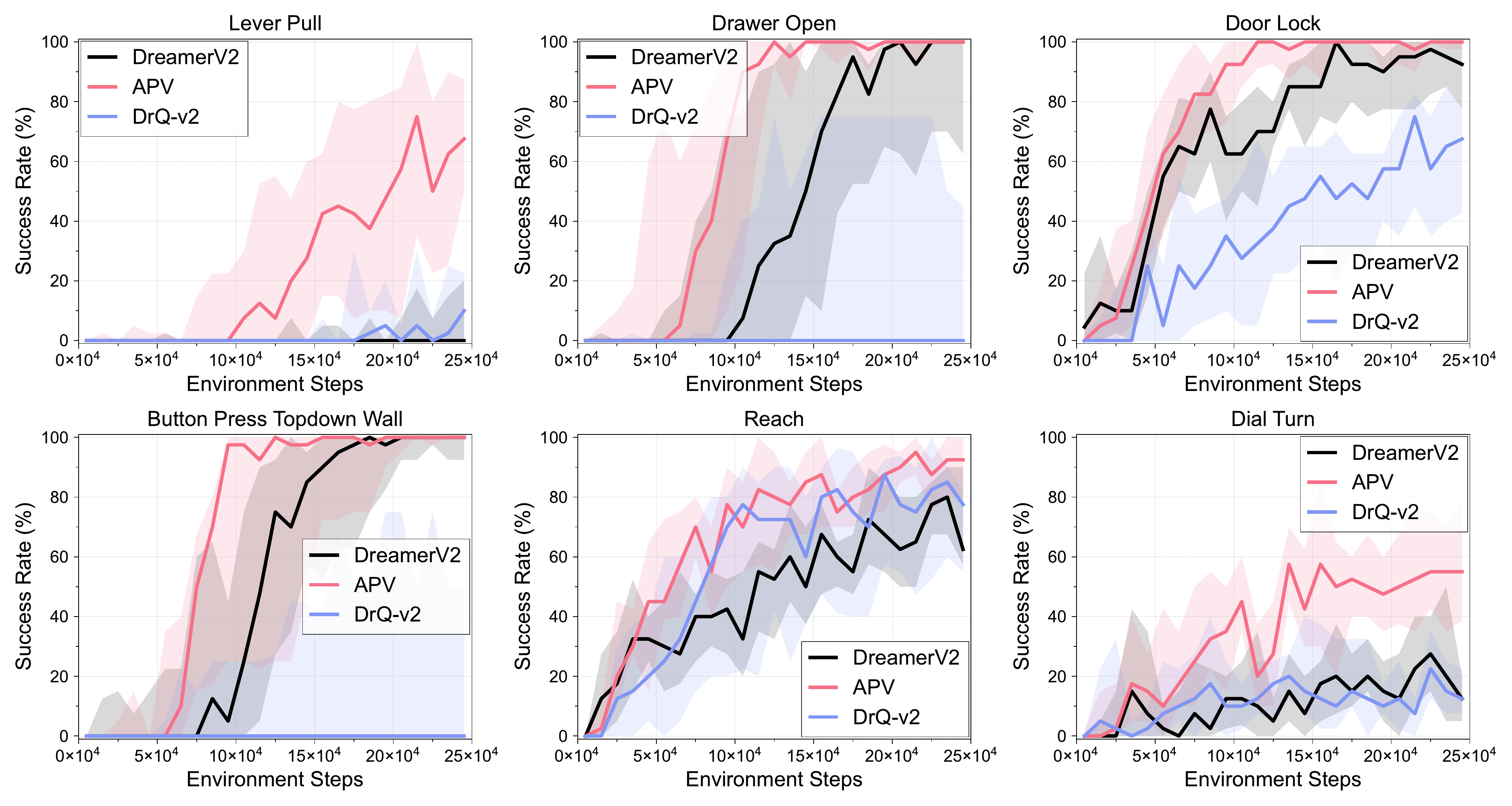}
\vspace{-0.2in}
\caption{
Learning curves on manipulation tasks from Meta-world as measured on the success rate.
The solid line and shaded regions represent the interquartile mean and bootstrap confidence intervals, respectively, across eight runs.}
\vspace{-0.1in}
\label{fig:metaworld_drq}
\end{figure*}

We report the performance of the state-of-the-art model-free RL method DrQ-v2 on vision-based robotic manipulation tasks from Meta-world.
We use image observations of $64 \times 64 \times 3$, and conduct the hyperparameter search over action repeat of $\{1, 2\}$ and the frame stacking of $\{3, 6\}$.
We find that action repeat of 1 with frame stacking of 6 performs the best.
As shown in \cref{fig:metaworld_drq}, we find that DrQ-v2 struggles to achieve competitive performance on most of the considered tasks, which necessitates more analysis to understand the reason of a failure.

\section{Real-World Video Prediction on Something-Something-V2}
\label{appendix:video_prediction_something}
We report the future frames predicted by our action-free video prediction model trained on real-world videos from Something-Something-V2~\citep{goyal2017something}.
One can see that the model severely suffers from underfitting, and generates blurry frames. Developing a lightweight, high-fidelity video prediction model for RL would be an interesting future direction.
For instance, we can consider adopting RSSM architecture based on Transformers~\citep{chen2022transdreamer}.

\begin{figure*} [h] \centering
\includegraphics[width=0.9\textwidth]{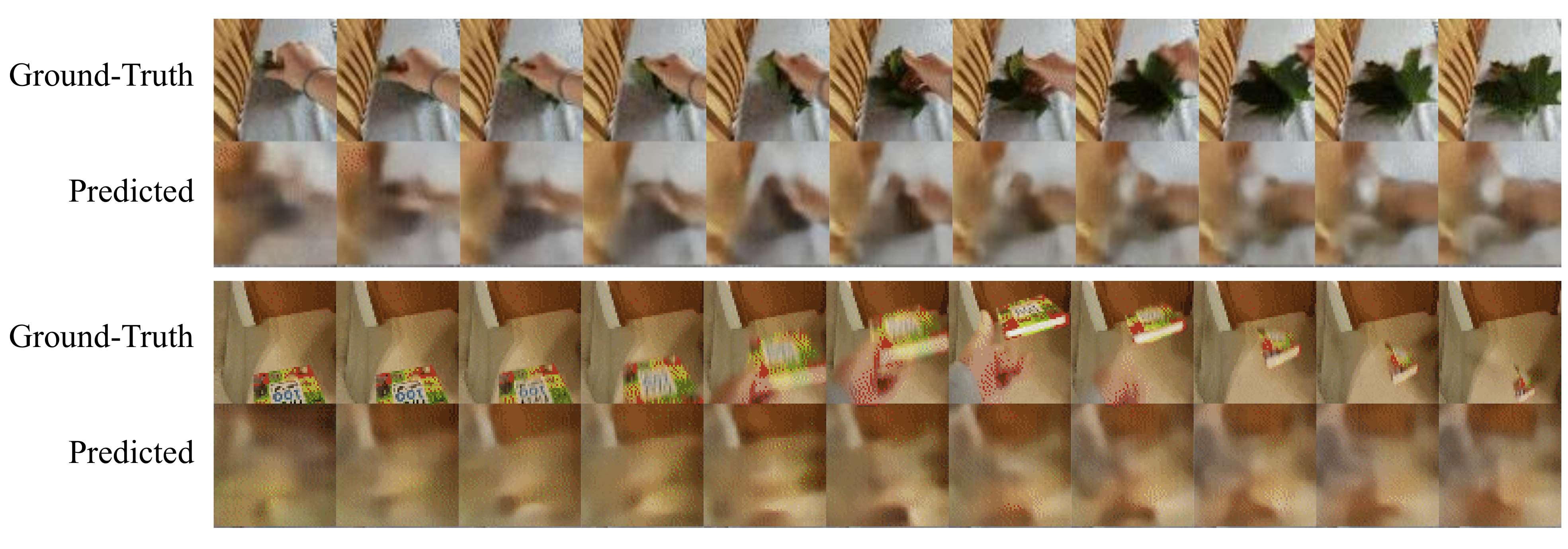}
\vspace{-0.2in}
\caption{
Future frames predicted by our action-free video prediction model on Something-Something-V2 dataset. We observe that our model severely suffers from underfitting, generating only blurry frames.}
\label{fig:video_prediction_something}
\end{figure*}

\section{Video Prediction on RLBench and Meta-world}
\label{appendix:video_prediction_simulated}
We report the future frames predicted by our action-free video prediction model trained on videos from RLBench~\citep{james2020rlbench} and Meta-world~\citep{yu2020meta}.
One can see that predicted frames on both datasets capture dynamics information of robots (e.g., how robots are moving towards objects), in contrast to the prediction on Something-Something-V2 where predicted frames are so blurry that it is difficult to see which behavior is performed in the videos.
We also observe that prediction quality in Meta-world is much better than that of RLBench, which is because Meta-world videos have more simple visuals and are collected in smaller number of tasks.

\begin{figure*} [h] \centering
\includegraphics[width=0.99\textwidth]{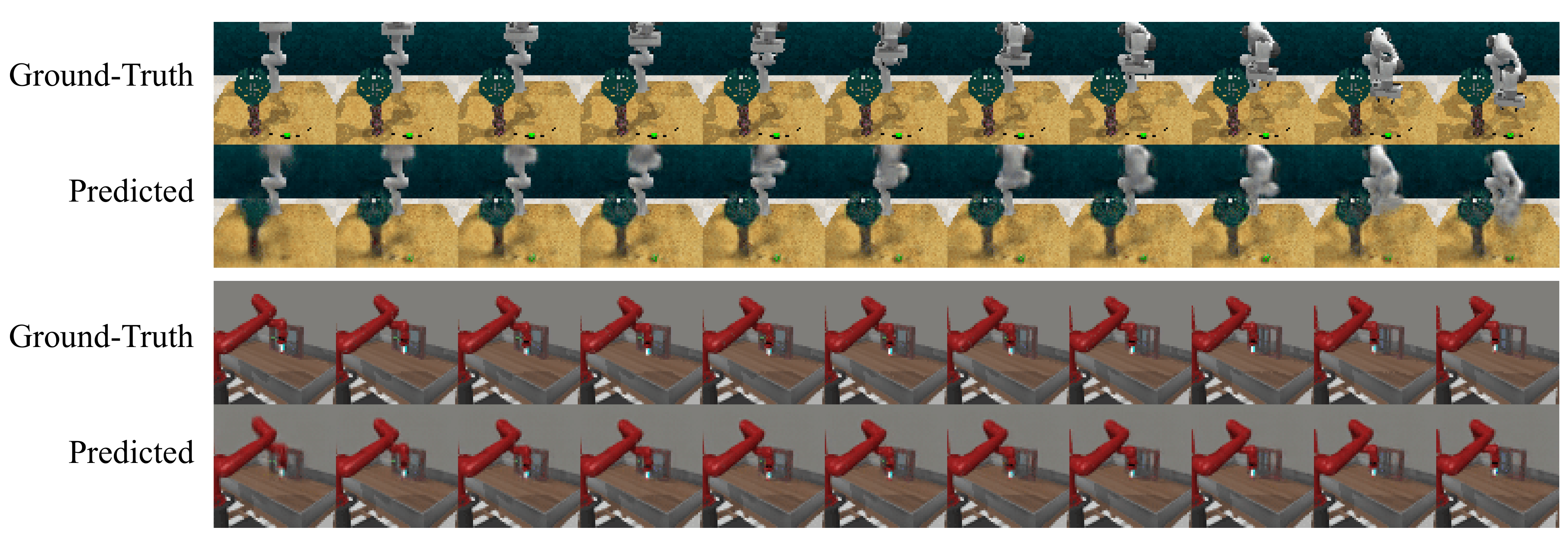}
\vspace{-0.2in}
\caption{
Future frames predicted by our action-free video prediction model on RLBench (top) and Meta-world (bottom). We observe that predicted frames on both datasets capture dynamics information of robots (e.g., how robots are moving towards objects), while prediction in RLBench is not high-quality as in Meta-world which is a more simple domain.}
\label{fig:video_prediction_simulated}
\end{figure*}

\section{Ablation Study on DeepMind Control Suite}
\label{appendix:dmc_ablation}
\begin{figure*} [h] \centering
\includegraphics[width=0.99\textwidth]{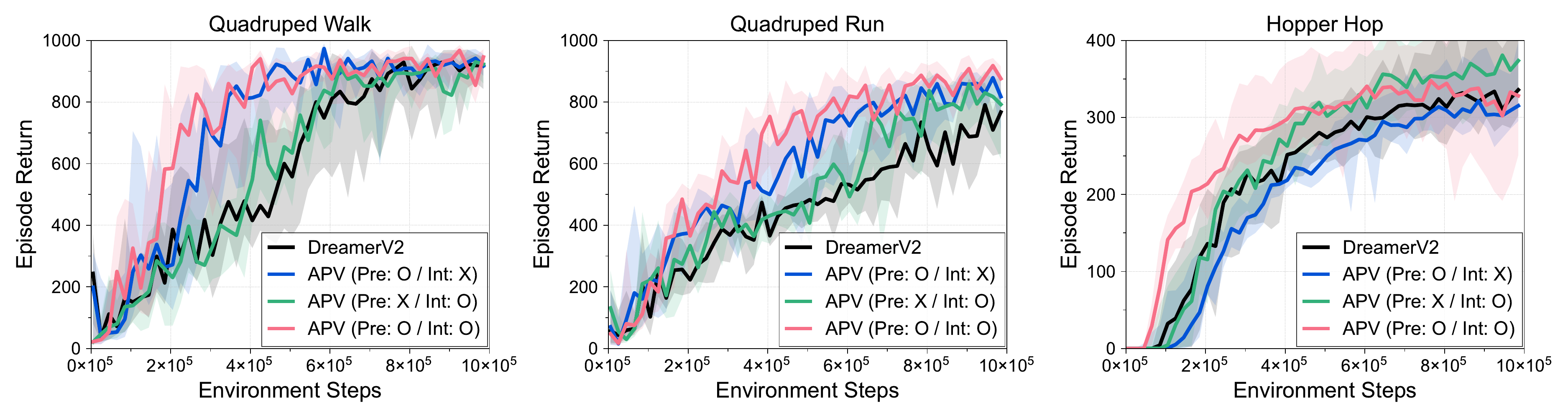}
\vspace{-0.2in}
\caption{
Learning curves of \ALG on locomotion tasks from DeepMind Control Suite as measured on the episode return.
The solid line and shaded regions represent the interquartile mean and bootstrap confidence intervals, respectively, across eight runs.}
\vspace{-0.1in}
\label{fig:metaworld_dmc_ablation_all}
\end{figure*}

We provide the results from ablation studies on locomotion tasks from DeepMind Control Suite in \cref{fig:metaworld_dmc_ablation_all}.
Interestingly, we find that pre-training can significantly improve the performance without intrinsic bonus on Quadruped tasks, but \ALG with only intrinsic bonus does not make significant difference over vanilla DreamerV2.
This might be because the visual observations in Quadruped tasks are very complex, so intrinsic bonus based on randomly initialized representations becomes not particularly useful in the tasks.
But \ALG with both pre-training and intrinsic bonus performs best, which shows that pre-training representations can provide useful information from the beginning of the fine-tuning.
On the other hand, on Hopper Hop, we find that \ALG without intrinsic reward struggles to outperform DreamerV2, which is because Hopper Hop is more difficult in terms of exploration.
We also observe that \ALG without pre-training cannot also outperform DreamerV2 before 300K environment steps, since the model needs a large amount of samples to learn representations that are useful for capturing the dynamics information of the environment.
By utilizing the dynamics information captured in the pre-trained representations for exploration, \ALG with both pre-training and intrinsic bonus performs best from the initial phase of fine-tuning.

\end{document}